%% file: main.tex
\documentclass[10pt,twocolumn,letterpaper]{article}
\usepackage[accsupp]{axessibility}  
\usepackage[table,dvipsnames]{xcolor}
\usepackage[pagenumbers]{cvpr} 
\input{preamble}

\definecolor{cvprblue}{rgb}{0.21,0.49,0.74}
\usepackage[pagebackref,breaklinks,colorlinks,citecolor=cvprblue]{hyperref}
\usepackage{makecell} 
\usepackage{multirow}
\usepackage{algorithm}
\usepackage{algorithmic}
\definecolor{lightgray}{gray}{0.9} 

\newcommand{\model}{Video-3D LLM}

\title{Video-3D LLM: Learning Position-Aware Video Representation for 3D Scene Understanding}

\def\authorBlock{
    Duo Zheng\thanks{Equal contribution.} \qquad
    Shijia Huang\footnotemark[1] \qquad
    Liwei Wang\thanks{Corresponding author.} \qquad \\
    The Chinese University of Hong Kong \\
    {\tt\small \{dzheng23, sjhuang, lwwang\}@cse.cuhk.edu.hk}
}
\author{\authorBlock}

\begin{document}
\maketitle
\input{sec/0_abstract}    
\input{sec/1_intro}

\input{sec/2_related}

\input{sec/3_method}
\input{sec/4_experiment}

\input{sec/5_conclusion}

{
    \small
    \bibliographystyle{ieeenat_fullname}
    \bibliography{main}
}

\clearpage
\newpage
\appendix 
\section*{Appendix}
\input{sec/X_appendix}

\end{document}

%% file: preamble.tex
\newcommand{\red}[1]{{\color{black}#1}}


%% file: sec/0_abstract.tex
\begin{abstract}

The rapid advancement of Multimodal Large Language Models (MLLMs) has significantly impacted various multimodal tasks. However, these models face challenges in tasks that require spatial understanding within 3D environments. Efforts to enhance MLLMs, such as incorporating point cloud features, have been made, yet a considerable gap remains between the models' learned representations and the inherent complexity of 3D scenes. This discrepancy largely stems from the training of MLLMs on predominantly 2D data, which restricts their effectiveness in comprehending 3D spaces. 
To address this issue, in this paper, we propose a novel generalist model, i.e., Video-3D LLM, for 3D scene understanding. By treating 3D scenes as dynamic videos and incorporating 3D position encoding into these representations, our Video-3D LLM aligns video representations with real-world spatial contexts more accurately. In addition, we have implemented a maximum coverage sampling technique to optimize the trade-off between computational cost and performance. Extensive experiments demonstrate that our model achieves state-of-the-art performance on several 3D scene understanding benchmarks, including ScanRefer, Multi3DRefer, Scan2Cap, ScanQA, and SQA3D.
Our code is available at \url{https://github.com/LaVi-Lab/Video-3D-LLM}.

\end{abstract}

%% file: sec/1_intro.tex
\section{Introduction}
\label{sec:intro}

The rapid development of Multimodal Large Language Models (MLLMs) \cite{blip2, gpt-4, llava, llava-onevision, qwen2vl, internvl, gemini, llama3, oryx, llavavideo} has demonstrated substantial capabilities in various multi-modal tasks, attracting significant attention from both academia and industry sectors. However, despite these advancements, recent studies \cite{coarse-correspondence, openeqa, msqa, spatial-vlm} indicate that current MLLMs face challenges when addressing tasks that necessitate spatial understanding and reasoning in 3D environments.

\begin{figure}[] 
\centering 
\includegraphics[width=0.49\textwidth]{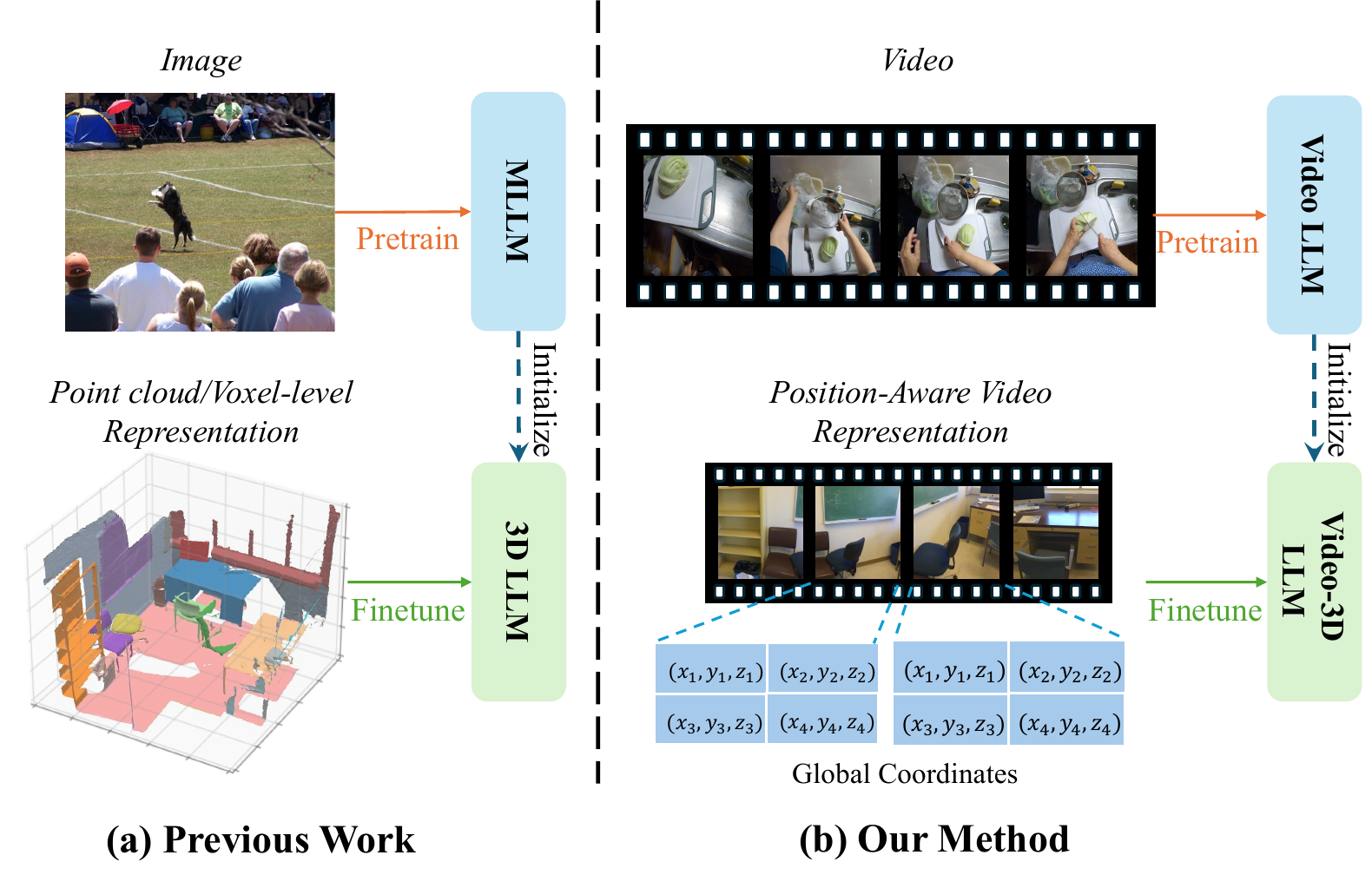} 
\caption{
Comparison of previous work and our method: (a) Previous 3D LLMs are initialized on MLLMs trained solely on image-text pairs, and learn point cloud or voxel representations via fine-tuning on 3D scenes. The 3D point clouds are reconstructed from RGB-D videos.
(b) Our method directly utilizes video frames and 3D coordinates as input, where the 3D coordinates are converted from depths through coordinate transformation.
We then transfer the ability of video understanding to 3D scene understanding by injecting position information into video representations.
}
\vspace{-10pt}
\label{fig:intro}
\end{figure}

Recent studies \cite{3d-llm, ll3da, pointllm, grounded-3dllm, llava3d, chatscene, chat3d, scenellm, robin3d} have focused on adapting MLLMs for enhanced 3D scene understanding. As depicted in Figure \ref{fig:intro} (a), these approaches develop comprehensive 3D scene-level representations using a variety of techniques. These include harnessing features from point clouds \cite{pointllm, grounded-3dllm, ll3da}, 
lifting multi-view image features to 3D space
~\cite{3d-llm, llava3d, scenellm}, and exploiting characteristics from recognized objects \cite{leo, chat3d, chatscene, robin3d}.


Although significant advancements have been achieved, a noticeable gap exists between the representations learned by MLLMs and the complexity of 3D scenes. This gap stems from the fact that MLLMs are primarily trained on fundamentally different data types, namely 2D images. While it is possible to further finetune MLLMs with 3D data, such as point clouds or voxels, the limited availability of labeled 3D scene data poses a challenge. 
\red{Consequently, the 2D visual knowledge embedded in MLLMs fails to fully unleash its potential in understanding 3D environments.}



In parallel, the abundance of video data has spurred interest in adapting Video LLMs to different domains,
\eg, 3D question answering \cite{oryx, llava-onevision, coarse-correspondence} and robotic manipulation \cite{rt2, openvla, gr1}. These methods benefit from extensive internet video datasets and pre-trained video models, revealing the immense potential for extending video modality to 3D modeling. However, as early attempts, they are still far from creating a model capable of handling diverse 3D tasks. Moreover, the absence of integrated spatial information—such as 3D locations and spatial relationships—in video representations constrains their capability to fully comprehend the 3D physical world. For instance, tasks that require an intricate understanding of 3D spatial relationships cannot rely solely on RGB data. This limitation underscores the necessity for incorporating more comprehensive spatial modeling into Video LLMs to enhance their effectiveness in 3D applications.

Therefore, in this paper, we propose a generalist model for 3D scene understanding, namely \model{}. 
\red{Our primary motivation is to effectively leverage the spatiotemporal priors inherent in Video LLMs and advance this modeling approach to a variety of 3D scene understanding tasks.}
As shown in Figure \ref{fig:intro}, our model is based on a Video LLM framework \red{that} processes 3D videos, \ie, video frames accompanied by the corresponding 3D global coordinates.
\red{The frames are sampled from raw RGB videos and the 3D coordinates are obtained by backprojecting each pixel in the depth images\footnote{Indoor 3D datasets \cite{scannet, scannetpp, 3rscan} are captured as RGB-D streams and then reconstructed into 3D point clouds.}.} 
To establish the correspondence in visual appearance and position information, we learn position-aware video representations by injecting 3D global coordinates into video features. Specifically, we encode the coordinate to 3D position encoding and add it to the video representations, serving as the input for the Video LLM. 

Our model offers several significant advantages. Firstly, it aligns video representations with their real-world spatial contexts, thereby equipping our model to handle various 3D tasks such as 3D visual grounding, 3D dense captioning, and 3D question answering. Secondly, it maintains both temporal and spatial contextual information in the video data, which helps to reduce the discrepancy between the pre-training data and actual 3D scenes. Additionally, we have developed a maximum coverage sampling strategy for frame selection. This approach views frame selection as a maximum coverage problem and adopts a greedy algorithm for its resolution. This strategy ensures the selection of the most informative frames, thus improving the model's capacity to discern diverse and essential spatio-temporal features within the video, while also ensuring efficient inference performance.

Our approach is to train a single model in a multi-task manner on varying 3D scene understanding tasks, including 3D question answering, 3D dense captioning, and 3D visual grounding.
Extensive experiments demonstrate that our Video-3D LLM achieves state-of-the-art performance on five 3D scene understanding benchmarks, \ie, ScanRefer \cite{scanrefer}, Multi3DRefer \cite{multi3drefer}, Scan2Cap \cite{scan2cap}, ScanQA \cite{scanqa} and SQA3D \cite{sqa3d}.
Notably, our method surpasses the previous state-of-the-art LLaVA-3D \cite{llava3d} by using only 26\% of its 3D data (223k vs. 859k), achieving improvements of \textbf{4.1\%} Acc@0.25 on ScanRefer, \textbf{4.6} CIDEr@0.5IoU on Scan2Cap, \textbf{2.9\%} EM on ScanQA, and \textbf{3.0\%} EM on SQA3D.
The impressive performance reveals the immense potential for adapting video models to 3D modality, establishing a new paradigm in 3D scene understanding.


%% file: sec/2_related.tex
\section{Related Work}

\noindent \textbf{LLMs for 3D Scene Understanding.} Recently, there has been a growing interest in integrating 3D information in LLMs \cite{3d-llm, grounded-3dllm, chat3d, leo, pointllm, chatscene, llava3d, ll3da, robin3d}, which advances the progress of 3D scene understanding. 3D-LLM \cite{3d-llm} introduces the LLM-based model for 3D physical world, which takes 3D features from rendered 2D images as input. PointLLM \cite{pointllm} utilizes a point encoder with a strong LLM for point cloud understanding.
LL3DA \cite{ll3da} leverages a Q-former to extract useful information from point cloud features, endowing humans with the capability to interact in 3D environments.
Grounded 3D-LLM \cite{grounded-3dllm} further introduces a projection module based on 3D detectors, which allows for generating object proposals from point-level features. Chat3D\cite{chat3d}, LEO \cite{leo}, and ChatScene \cite{chatscene} take use of off-the-shelf 3D detectors for proposal generation, and then incorporate the object-centric representations into LLMs.
LLaVA-3D \cite{llava3d} introduces 3D-patch representations, which aggregate 2D-patch features in voxel space. 
Robin3D \cite{robin3d} tries to enhance 3D scene understanding via data generation.
It is important to note that existing 3D Large Language Models (3D LLMs) typically transform 3D scenes into voxel-level or point cloud-level 3D representations as input for modeling purposes. However, these approaches create a disconnect with pre-trained multi-modal large language models (MLLMs), which are primarily trained on extensive 2D datasets, such as images, and only fine-tuned on a limited amount of 3D scene data. To address this challenge, our method incorporates 3D information (e.g., coordinates in Figure \ref{fig:intro} (b)) into a new video representation. This enhancement maximizes the use of pre-trained 2D Video LLMs, leveraging their full potential.

\vspace{3pt}
\noindent \textbf{Video-Language Models for 3D Understanding.}
We have witnessed the rapid development of Video LLMs \cite{videollama, llavavideo, oryx, mvbench, vila}. There is also a trend of leveraging Video LLMs for 3D tasks, including 3D question answering \cite{oryx, llava-onevision, coarse-correspondence} and robotic manipulation \cite{rt2, openvla, gr1}. LLaVA-OneVision \cite{llava-onevision} and Oryx MLLM \cite{oryx} incorporate 3D question-answering datasets into instruction data, which deliver competent results on 3D question-answering tasks. However, these models do not capture detailed 3D spatial information, which limits their performance in addressing other 3D tasks that require precise spatial alignments. Furthermore, recent work \cite{coarse-correspondence} has emphasized the importance of identifying key object correspondences across frames through visual prompting. In contrast, our approach directly incorporates 3D positional information into video representation learning, which enhances the capability to tackle more complex tasks that demand a thorough spatial understanding of the 3D environment.

%% file: sec/3_method.tex
\begin{figure*}[t]{ 
\centering 
\includegraphics[width=0.93\textwidth]{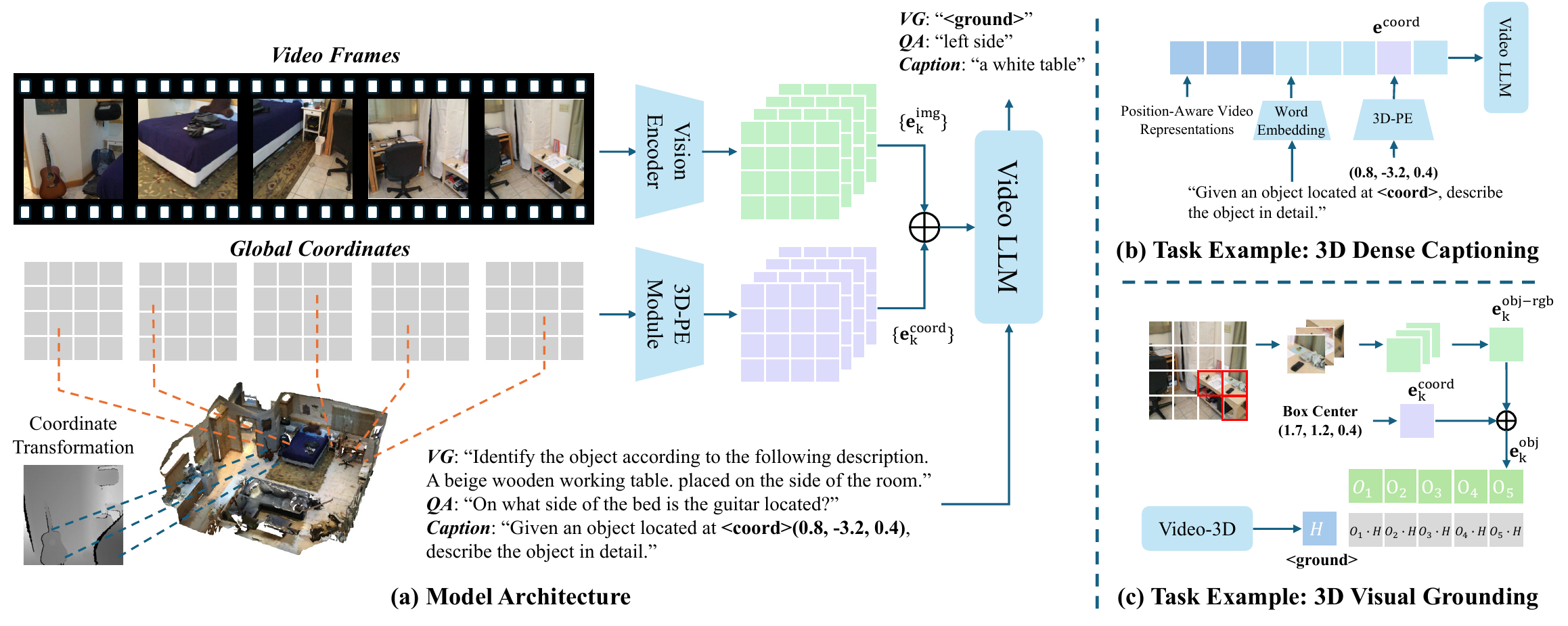} 
\vspace{-5pt}
\caption{
The overview of the model architecture.
(a) shows the integration of video sequence and global coordinates for creating position-aware video representations.
(b) and (c) detail the examples of 3D dense captioning and 3D visual grounding, respectively.
Our approach can generalize well to other 3D tasks.
}
\vspace{-10pt}
\label{fig:model}
}
\end{figure*}

\section{Method}
We propose a generalist model for 3D scene understanding, namely \model{}, which comprises a visual encoder, a 3D position encoding module, and a Video LLM backbone.
\red{
As shown in Figure \ref{fig:model}, the model takes the input as: 1) video streams captured from the 3D scene and 2) the associated 3D coordinate maps obtained through back-projection across all frames.
In contrast to prior work \cite{3d-llm, ll3da, llava3d} that converts video frames into explicit 3D representations (\eg, point clouds or voxels), our model directly processes temporal sequences of video frames, preserving both temporal and spatial contextual information in the video streams.
}

Given that entire frame sequences can be redundant, implementing an effective frame selection strategy is crucial, as it significantly influences both performance and computational efficiency. Subsequently, our goal is to enhance the Video LLM with position awareness \red{and adapt it to various 3D scene understanding tasks.} This is achieved by encoding spatial coordinates into the 3D position encoding (3D-PE) and integrating them into the video representation learning. 
In this section, we detail three key components of our approach: the frame sampling strategy (\ref{sec:frame_sampling}), position-aware video representation (\ref{sec:spatial_representation}), and the multi-task training (\ref{sec:training_obj}).

\subsection{Frame Sampling Strategy} \label{sec:frame_sampling}


\red{Representing} 3D scenes as video sequences presents two primary challenges: (1) Due to GPU memory constraints, the Video LLM can only process a limited number of frames at a time. This necessitates the sampling of a subset of frames from the extensive raw video sequence to manage resources effectively. (2) It is crucial for the video sequence to encompass as much of the entire 3D scene as possible, since any omission of scene content could result in a significant and irreversible decline in model performance. To address these challenges, we introduce a maximum coverage strategy for frame sampling. This approach involves preprocessing the selected frames \textit{offline} and applying the strategy consistently during both the training and inference phases to ensure comprehensive scene coverage and efficient memory usage.

\vspace{3pt}
\noindent \textbf{Frame Sampling as Maximum Coverage Problem.}
Given a raw RGB-D video, each frame captures a portion of the 3D scene. We aim to select fewest possible frames that maximize coverage of the 3D scene, which could be formulated as a maximum coverage problem.
Formally, let \( F = \{f_1, f_2, \ldots, f_n\} \) represent the set of all frames, and \( V = \{v_1, v_2, \ldots, v_m\} \) represent the set of all voxels in the 3D space. 
\red{Each frame \( f_k \) covers a subset of voxels \( V_k \subseteq V \). 
To identify the contained voxels of each frame, we first back-project each pixel from the depth image to global coordinates, then discretize these coordinates into voxel grids.
}
The objective is to find a subset of frames \( S \subseteq F \) such that the union of covered voxels \( \bigcup_{f_k \in S} V_k \) is maximized.

\vspace{3pt}
\noindent \textbf{Greedy Solution.}
Since the maximum coverage problem is NP-hard, we employ a greedy algorithm to solve it, which can obtain an approximation ratio of \(1-1/e\) \cite{KHULLER199939}.
As illustrated in the Algorithm \ref{alg:greedy}, the approach iteratively selects the frame with the largest increase in uncovered voxel coverage. Frames are added until the desired number is reached or the coverage ratio exceeds a predefined threshold.
This stop condition ensures a balance between computational efficiency and the coverage for varying scenes.

\begin{algorithm}
\small
\caption{Maximum Coverage Sampling}
\begin{algorithmic}[1]
\REQUIRE Set of frames \( F = \{f_1, f_2, \ldots, f_n\} \), voxel sets \( V_k \) for each frame \( f_k \), budget \( K \)
\ENSURE Subset \( S \subseteq F \) maximizing voxel coverage
\STATE Initialize \( S \leftarrow \emptyset \)
\STATE Initialize \( U \leftarrow \emptyset \) \COMMENT{Set of covered voxels}
\WHILE{size of \( S \) is less than \( K \)}
    \STATE Select \( f^* = \arg\max_{f_k \in F \setminus S} |V_k \setminus U| \)
    \STATE Add \( f^* \) to \( S \)
    \STATE Update \( U \leftarrow U \cup V_{f^*} \)
\IF{ Stop condition is met}
        \STATE $\mathbf{break}$
    \ENDIF
\ENDWHILE
\RETURN \( S \)
\end{algorithmic}
\label{alg:greedy}
\end{algorithm}
\vspace{-5pt}

\subsection{Position-Aware Video Representation} \label{sec:spatial_representation}
After completing frame sampling, we obtain a sequence of RGB frames, depth images, and the camera's intrinsic and extrinsic parameters. To create a position-aware video representation, we first transform the depth information into 3D coordinates within a global coordinate system. We then encode the visual embeddings along with the 3D position encodings (3D-PE) to enhance spatial awareness.

\vspace{5pt} \noindent \textbf{Camera Coordinate Transformation.}
Given a depth image $d_k \in \mathbb{R}^{H\times W}$, an extrinsic matrix $T_k \in \mathbb{R}^{4\times 4}$, and a camera intrinsic matrix $K$, \red{we can backproject each pixel at the position $(i, j)$ in the depth image to the global coordinates $c_k(i,j)$:}

\vspace{-0.5cm}
\begin{equation}
\small
\begin{bmatrix}
    c_k(i,j) \\
    1
\end{bmatrix} = 
T_k \cdot \begin{bmatrix}
d_k (i,j) \cdot K^{-1} \cdot \begin{bmatrix}
j \\
i \\
1
\end{bmatrix} \\
1
\end{bmatrix}.
\end{equation}
\red{We perform the above process for each sampling frame, resulting in a set of global coordinates \(\{c_k\}_{k=1}^l\) and their corresponding RGB images \(\{f_k\}_{k=1}^l\). Here, both \(c_k\) and \(f_k\) are in \(\mathbb{R}^{H \times W \times 3}\).}


\vspace{5pt} \noindent \textbf{Visual Embedding.}
We first encode each frame into visual embeddings via a Vision Transformer (ViT) \cite{vit}. In concrete, given a frame $f_i \in \mathbb{R}^{H \times W \times 3}$, the image will first be split into a series of patches at the patch size $P$, which are then fed into the ViT to produce visual embeddings $\mathbf{e}^{img}_k \in R^{H'\times W'\times d}$, where  \(H' = \left\lfloor \frac{H}{P} \right\rfloor\), \(W' = \left\lfloor \frac{W}{P} \right\rfloor\) and $d$ is the feature dimension.

\vspace{5pt} \noindent \textbf{3D Position Encoding.}
Since the video frame is divided into image patches, we need to pool the coordinate information of each image patch.
For each global coordinate map $c_k\in R^{H\times W \times 3}$, we divide the coordinates into patches identical to those of the image patches.
Subsequently, we compute the average coordinates for each patch with:
\begin{equation}
\small
    c_k'(i, j) = \frac{1}{P^2} \sum_{(u, v) \in \mathcal{P}(i,j)} c_k(u, v),
\end{equation}
\red{where \(\mathcal{P}(i,j)\) denotes the patch region corresponding to position \((i,j)\) and $c_k' \in R^{H'\times W'\times d}$ are the average coordinates for \(\mathcal{P}(i,j)\).}
Due to the small size of the patches, the averaged coordinates retain sufficiently precise positional information. We also explored alternative coordinate pooling methods in our ablation study.

We adopt sinusoidal position encoding \cite{DBLP:conf/nips/VaswaniSPUJGKP17} to encode the coordinates. For 3D coordinates $(x, y, z)$, we first map the coordinate onto a discrete grid. The encoding for the $x$ coordinate is then defined as:
\begin{align}
\small
\text{PE}(x, 2i) &= \sin\left(\frac{x}{10000^{2i/\left\lfloor \frac{d}{3} \right\rfloor }}\right), \\
\text{PE}(x, 2i+1) &= \cos\left(\frac{x}{10000^{2i/\left\lfloor \frac{d}{3} \right\rfloor }}\right).
\end{align}

\noindent Similar calculations are applied to the $y$ and $z$ coordinates. 
The PE of $(x,y,z)$ are concatenated to obtain the final coordinate embeddings $\mathbf{e}^{\mathrm{coord}}_k \in R^{H' \times W' \times d}$.
Lastly, the coordinate embeddings are added to the visual embeddings to form the position-aware video representations, denoted by:
\begin{equation}
\small
    \mathbf{e}^{vis}_k = \mathbf{e}^{img}_k + \mathbf{e}^{coord}_k.
\end{equation}

\subsection{Multi-Task Training}\label{sec:training_obj}
Our approach is to build a generalist model capable of handling multiple tasks with the single learned model. To achieve this, we train our model on a diverse, multi-task dataset that encompasses various 3D scene understanding tasks.
During training, we randomly sample a single task type for each batch and train exclusively on data specific to that task. 
For general 3D scene understanding tasks, such as 3D question answering and 3D dense captioning, we use cross-entropy loss to supervise text generation.
For the 3D visual grounding task, to locate more accurately, we use the designed 3D visual grounding loss to supervise the 3D proposal selection.

\vspace{5pt} \noindent \textbf{Cross-Entropy Loss.}
Given a position-aware video representation and a textual instruction, the language modeling objective aims to optimize the cross-entropy loss $\mathcal{L}_{\mathrm{CE}}$:
\begin{equation}
\small
    \mathcal{L}_{\mathrm{CE}} = - \sum \log(y | \{\mathbf{e}_k^{vis}\}_{k=1}^l, \{\mathbf{e}_k^{text}\}_{k=1}^q),
\end{equation}
\noindent where $y$ is the ground truth response, $\{\mathbf{e}_k^{vis}\}_{k=1}^l$ are position-aware video representations for the video and $\{\mathbf{e}_k^{text}\}_{k=1}^q$ are text embeddings.

\red{For the dense captioning task, we ask the model to describe objects based on their center coordinates.}
As shown in Fig~\ref{fig:model} (b), we obtain the 3D position encoding of the bounding box center in the same manner as described in Sec~\ref{sec:spatial_representation}. This position encoding is added to the embedding of the special token \(\langle \text{coord} \rangle\) to provide location information.

\input{table/main_result}

\vspace{5pt}
\noindent \textbf{3D Visual Grounding Objective.}
Previous studies~\cite{3d-llm, scanreason} have 
demonstrated that directly outputting 3D bounding boxes is quite challenging for LLM. 
\red{To enable our model to perform 3D visual grounding, we follow previous work \cite{chatscene, mvt, 3dvista} to model the task as a proposal classification task--selecting the target objects from a list of detected proposals.} 
As illustrated in Fig~\ref{fig:model} (c), given a list of object proposals, we extract object features for each object from the visual embeddings. 
Specifically, for each object \(b_k\), we check each patch to see if more than 50\% of its points are contained within \(b_k\), and then apply average pooling to the features of all selected patches, to obtain the 2D object features \(\mathbf{e}^{\text{obj-rgb}}_k\).
Lastly, we add the 3D position encoding of the center coordinate \(\mathbf{e}^{\text{obj-coord}}_i\) with \(\mathbf{e}^{\text{obj-rgb}}_i\) to obtain the object representation \(\mathbf{e}^{\text{obj}}_i\).
During training, we utilize InfoNCE loss \cite{DBLP:journals/corr/abs-1807-03748} to optimize the similarity between the ground truth object feature and the hidden states $h$ of the \(\langle \text{ground} \rangle\) token:
\begin{equation}
\small
    \mathcal{L}_{Grd} = \frac{
    \sum_{k\in O^+}\exp(f(\mathbf{e}^{\text{obj}}_{k})\cdot g(h) / \tau)
    }{\sum_{k\in O} \exp(f(\mathbf{e}^{\text{obj}}_{k})\cdot g(h) / \tau)},
\end{equation}
\vspace{-0.2cm}

\noindent \red{where \(O^+\) and \(O\) respectively represent the sets of positive objects and all objects}, $f$ and $g$ are two-layer learnable MLPs, and $\tau$ is the temperature. 






%% file: table/main_result.tex
\begin{table*}[]{
\centering
\small
\resizebox{0.98\linewidth}{!}{
\setlength{\tabcolsep}{2.5mm}{
\begin{tabular}{lcccccccccc}
\toprule 
\multirow{2}{*}{Method} & \multirow{2}{*}{\shortstack{3D\\Generalist}} & \multicolumn{2}{c}{ScanRefer} & \multicolumn{2}{c}{Multi3DRef} & \multicolumn{2}{c}{Scan2Cap}  & \multicolumn{2}{c}{ScanQA} & \multicolumn{1}{c}{SQA3D} \\ 
\cmidrule(lr){3-4} \cmidrule(lr){5-6} \cmidrule(lr){7-8} \cmidrule(lr){9-10} \cmidrule(lr){11-11} 
 & & \scalebox{0.85}[1]{Acc@0.25} & \scalebox{0.85}[1]{Acc@0.5} & \scalebox{0.9}[1]{F1@0.25} & \scalebox{0.9}[1]{F1@0.5} & \scalebox{0.85}[1]{B-4@0.5} & \scalebox{0.85}[1]{C@0.5} & C & EM & EM \\ \midrule
  \multicolumn{5}{l}{\textit{\textbf{Expert Models}}}\\  \addlinespace[0.1cm]
{\color[HTML]{969696} ScanRefer~\cite{scanrefer}} & & {\color[HTML]{969696} 37.3} & {\color[HTML]{969696} 24.3} \\
{\color[HTML]{969696} MVT~\cite{mvt}} & & {\color[HTML]{969696} 40.8} & {\color[HTML]{969696} 33.3} \\
{\color[HTML]{969696} 3DVG-Trans~\cite{3dvg-trans}} & & {\color[HTML]{969696} 45.9} & {\color[HTML]{969696} 34.5} \\
{\color[HTML]{969696} ViL3DRel~\cite{vil3drel}} & & {\color[HTML]{969696} 47.9} & {\color[HTML]{969696} 37.7} \\
{\color[HTML]{969696} M3DRef-CLIP~\cite{multi3drefer}} & & {\color[HTML]{969696} 51.9} & {\color[HTML]{969696} 44.7} & {\color[HTML]{969696} 42.8} & & {\color[HTML]{969696} 38.4} \\
{\color[HTML]{969696} Scan2Cap~\cite{scan2cap}} & & & & &  & {\color[HTML]{969696} 22.4}& {\color[HTML]{969696} 35.2} \\
{\color[HTML]{969696} ScanQA~\cite{scanqa}} & & & & & & &  & {\color[HTML]{969696} 64.9}  &{\color[HTML]{969696} 21.1}&  {\color[HTML]{969696} 47.2}  \\
{\color[HTML]{969696} 3D-VisTA~\cite{3dvista}} & & {\color[HTML]{969696} 50.6} & {\color[HTML]{969696} 45.8} & &  & {\color[HTML]{969696} 34.0}& {\color[HTML]{969696} 66.9} &  {\color[HTML]{969696} 69.6}  & {\color[HTML]{969696} 22.4}& {\color[HTML]{969696} 48.5} \\ 
 \midrule
 \multicolumn{5}{l}{\textit{\textbf{2D LLMs}}}\\  \addlinespace[0.1cm]
Oryx-34B~\cite{oryx} & & -- & -- & -- & -- & -- & -- & 72.3 & -- & -- \\
\rowcolor{lightgray} LLaVA-Video-7B \cite{llavavideo} & & -- & -- & -- & -- & -- & -- & 88.7 & -- & 48.5 \\
\midrule
\multicolumn{5}{l}{\textit{\textbf{3D LLMs}}}\\  \addlinespace[0.1cm]
\scalebox{0.9}[1]{3D-LLM(Flamingo)}\cite{3d-llm} & & 21.2 & -- & -- & -- & --& --  & 59.2 & 20.4 & --   \\
\scalebox{0.9}[1]{3D-LLM(BLIP2-flant5)}\cite{3d-llm} & & 30.3 & -- & -- & -- & -- & --  & 69.4 & 20.5 & --  \\
Chat-3D~\cite{chat3d} &  & -- & -- & -- & -- & -- & -- & 53.2 & --   \\
Chat-3D v2 ~\cite{chatscene} & \checkmark & 42.5 & 38.4 & 45.1 & 41.6 & 31.8  & 63.9 & 87.6 & -- & 54.7 \\
LL3DA~\cite{ll3da}& \checkmark  & -- & -- & -- & -- & 36.0 & 62.9  & 76.8 & -- & --  \\
SceneLLM \cite{scenellm} & \checkmark & -- & -- & -- & -- & -- & -- & 80.0 & 27.2 & 53.6 \\
LEO \cite{leo} & \checkmark & -- & -- & -- & --  & 38.2 & 72.4 & 101.4 & 21.5 & 50.0 \\
Grounded 3D-LLM \cite{grounded-3dllm} & \checkmark & 47.9 & 44.1 & 45.2 & 40.6 & 35.5 & 70.6 & 72.7 & -- & -- \\ 
PQ3D \cite{pq3d} & \checkmark & 57.0 & 51.2 & -- & 50.1 & 36.0 & 80.3 & -- & -- & 47.1 \\
ChatScene \cite{chatscene}& \checkmark  & 55.5 & 50.2 & 57.1 & 52.4 & 36.3 & 77.1 & 87.7 & 21.6 & 54.6 \\
LLaVA-3D \cite{llava3d}& \checkmark & 54.1 & 42.4 & -- & -- & 41.1 & 79.2 & 91.7 & 27.0 & 55.6 \\
\rowcolor{lightgray} \textbf{\model{} (MC)} & \checkmark & 57.9 & 51.2 & 57.9 & 52.4 & 40.2 & 80.0 & 100.5 & 29.5 & 57.7 \\
\rowcolor{lightgray} \textbf{\model{} (Uniform)} & \checkmark & \textbf{58.1} & \textbf{51.7} & \textbf{58.0} & \textbf{52.7} & \textbf{41.3}  & \textbf{83.8} & \textbf{102.1 }& \textbf{30.1}  & \textbf{58.6} \\
\bottomrule
\end{tabular}}
}
\vspace{-3pt}
\caption{Overall performance comparison. 
``Expert models'' are customized for specific tasks through task-oriented heads. 
``3D Generalist'' means the model can perform multiple 3D tasks in a single model.
LLaVA-Video is assessed in a zero-shot setting.
} 
\label{tab: 3d llm performance}
\vspace{-10pt}
}
\end{table*}

%% file: sec/4_experiment.tex
\section{Experiments}
In this section, we first compare the overall performance of \model{} with top-tier models and also investigate the effectiveness of all components.

\subsection{Experimental Setup}
\noindent \textbf{Datasets.}
We conduct experiments across five 3D scene understanding benchmarks.
For visual grounding, we test our model on ScanRefer \cite{scanrefer} and Multi3DRefer \cite{multi3drefer}, which require localizing objects in single-target and multiple-target scenarios, respectively.
For dense captioning, we utilize the Scan2Cap \cite{scan2cap} benchmark, which involves densely generating descriptions for all objects in 3D scenes.
For question answering, we use the ScanQA \cite{scanqa} for spatial reasoning and SQA3D \cite{sqa3d} for situated reasoning.
All these datasets are sourced from the ScanNet \cite{scannet}, a richly annotated RGB-D video dataset containing 1,513 scans in 3D scenes.
We pre-process video frames for each scan at 3 FPS and extract the corresponding extrinsic and camera intrinsic parameters.
For evaluation, we follow previous work \cite{chatscene, llava3d, grounded-3dllm} to adopt the validation sets for ScanRefer, Multi3DRefer, Scan2Cap, and ScanQA, and adopt the test set for SQA3D.

\vspace{5pt}
\noindent \textbf{Metrics.}
We adopt widely used evaluation metrics for each of these benchmarks. 
For ScanRefer \cite{scanrefer}, we report thresholded accuracy metrics, specifically Acc@0.25 and Acc@0.5, where a prediction is considered correct if its Intersection over Union (IoU) with the ground truth exceeds 0.25 and 0.5, respectively. 
For Multi3DRefer \cite{multi3drefer}, which involves grounding a variable number of target objects, we use the F1 score at IoU thresholds of 0.25 and 0.5. 
For Scan2Cap \cite{scan2cap}, we apply CIDEr@0.5IoU and BLEU-4@0.5IoU (denoted as C@0.5 and B-4@0.5), combining traditional image captioning metrics with IoU between predicted and reference bounding boxes. 
For ScanQA \cite{scanqa}, we use CIDEr \cite{cider} and exact match accuracy, referred to as C and EM, respectively. 
Finally, for SQA3D \cite{sqa3d}, we evaluate performance using exact match accuracy (EM).

\vspace{5pt}
\noindent \textbf{Implementation Details.}
We build \model{} based on the LLaVA-Video 7B \cite{llavavideo}, an open-sourced video LLM based on the QWen2-7B \cite{qwen2}.
We use the Adam optimizer to train our model for one epoch with a batch size of 16 and a warmup ratio of 0.03.
During the warmup phase, the learning rates peak at 1e-5 for the LLM and 2e-6 for the vision encoder. 
All experiments are conducted on 8 A100-80G GPUs.
For 3D visual grounding and dense captioning, in training, we use the ground truth objects as the candidates. While in inference, we follow \cite{chatscene, leo} to employ Mask3D \cite{mask3d} to generate object proposals.
The temperature $\tau$ for InfoNCE loss is 0.07.

\subsection{Comparison with State-of-the-art Methods}

\subsubsection{Comparison Baselines}
For a comprehensive comparison, we include both expert models designed for specific tasks and LLM-based models.

\vspace{5pt}
\noindent \textbf{{Expert Models.}}
For ScanRefer \cite{scanrefer}, we compare our method with ScanRefer \cite{scanrefer}, MVT \cite{mvt}, 3DVG-Trans \cite{3dvg-trans}, ViL3DRel \cite{vil3drel}. M3DRef-CLIP \cite{multi3drefer} further extends 3D grounding capabilities to multi-target scenarios. Scan2Cap \cite{scan2cap} and ScanQA \cite{scanqa} provide initial benchmarks for the Scan2Cap and ScanQA datasets, respectively. 3D-VisTA \cite{3dvista} is pre-trained on large-scale scene-text pairs and then finetuned on specific tasks.

\vspace{5pt}
\noindent \textbf{{2D LLMs.} }
Oryx \cite{oryx} has included the ScanQA dataset in its training stage.
We also test the zero-shot performance of LLaVA-Video \cite{llavavideo}.

\vspace{5pt}
\noindent \textbf{{3D LLMs.}}
3D-LLM \cite{3d-llm} is the first LLM-based model for 3D scene undersanding.
SceneLLM and LL3DA \cite{ll3da} enrich the 3D representations with point cloud features.
Chat3D \cite{chat3d}, LEO \cite{leo}, and ChatScene \cite{chatscene} incorporate object representations into 3D LLMs.
Grounded 3D-LLM \cite{grounded-3dllm}, PQ3D \cite{pq3d} and LLaVA-3D \cite{llava3d} deliver impressive results on 3D visual grounding by co-training a 3D detector.
All these methods apply 3D point cloud features or projecting multi-view image features into 3D space, while ours directly works on the video representations.

\subsubsection{Results}

\input{table/ablation_frame}

We present the overall comparison with leading methods in Table \ref{tab: 3d llm performance}.
``\model{} (Uniform)'' is trained using uniform sampling with 32 frames, while ``\model{} (MC)'' is trained using maximum coverage sampling with a coverage ratio of 95\% and a maximum frame number of 32.
``\model{} (Uniform)'' achieves state-of-the-art performance on a variety of tasks including 3D visual grounding, 3D dense captioning, and 3D question answering, while ``\model{} (MC)'' delivers similar results with only half the inference time (527ms vs. 1050 ms).

\vspace{5pt}
\noindent \textbf{3D Visual Grounding.}
For 3D visual grounding, we follow previous work \cite{mvt, 3dvista, chatscene} to detect all objects and then make predictions over the object proposals\footnote{For a fair comparison, we use the Mask3D-generated object proposals provided by LEO \cite{leo} for both 3D visual grounding and dense captioning.\label{fn:object_proposals}}.
Our model achieves the highest accuracy, with Acc@0.25 at 58.1\% and Acc@0.5 at 51.7\% on ScanRefer, and F1@0.25 at 58.0\% and F1@0.5 at 52.7\% on Multi3DRefer.
As previous 3D LLMs either use detected object proposals (\eg, ChatScene, Chat3D) or train an additional grounding module based on a 3D detector (\eg, Grounded 3d-LLM, PQ3D, LLaVA-3D), we can compare with these 3D LLMs fairly.
Specifically, our model improves Acc@0.25 by 2.6\% on ScanRefer and F1@0.25 by 0.9\% on Multi3DRefer compared to ChatScene, which uses the same object proposal as ours.

\vspace{5pt}
\noindent \textbf{3D Dense Captioning.}
Following the previous setting \cite{leo, chatscene}, we generate captions for each detected object proposal\footref{fn:object_proposals}.
Our method demonstrates superior performance in Scan2Cap, achieving 41.3 at B-4@0.5 and 83.8 at C@0.5.
The results reveal that our method connects video content with its position information by injecting 3D-PE into video representations.

\vspace{5pt}
\noindent \textbf{3D Question Answering.}
Our model outperforms the best competitors with 30.1\% EM on ScanQA and 58.6\% EM on SQA3D, which could be attributed to the strong representations inherited from Video LLMs.
Existing Video LLMs achieve competent results in a zero-shot manner, suggesting that current 3D QA tasks may not sufficiently address the challenges of spatial reasoning in 3D scenes.

\subsection{Ablation Study}

\input{table/ablation_pe}

\noindent \textbf{Effectiveness of Frame Sampling.}
In Table~\ref{table:ablation_frame}, we assess the effectiveness of the frame sampling strategy across varying numbers of frames. To evaluate the model's efficiency, we calculated the average inference speed on ScanQA.
In the fixed frame number setting, we sample frames until reaching the desired number. As shown in the table, the model's performance improves with an increasing number of frames, though inference time also rises. \model{} surpasses many previous methods with only 8 frames, demonstrating that our position-aware video paradigm can effectively model 3D scene understanding tasks.
With 8 and 16 frames, maximum coverage sampling (``MC'') significantly enhances performance for all tasks by capturing a more complete 3D scene. Notably, with 8 frames, it improves Acc@0.25 by $4.54$ on ScanRefer and C@0.5 by $4.26$ on Scan2Cap compared to the uniform sampling strategy (``Uniform''). 
With 32 frames, the ``Uniform'' strategy shows comparable results with ``MC'', as this number of frames is more than enough to cover the 3D scene.
In the adaptive frame number setting, frames are sampled until over 95\% of the scene's voxels are covered or a maximum of 32 frames is reached. This allows for flexible adjustment based on scene size. As shown in the results, ``MC$^{*}$'' uses an average of $18$ frames across all scenes, achieving similar performance to the 32-frame uniform strategy while offering better inference speed. Meanwhile, compared to LLaVA-3D~\cite{llava3d}, ``MC$^{*}$'' shows superior performance while achieving similar inference speed.
\red{\textit{Maximum coverage sampling is performed once per scene, and the average time spent on ScanQA is only 17.8ms per question, which is negligible compared to the inference time}.}

\begin{figure*}[htbp] 
\centering 
\includegraphics[width=0.94\textwidth]{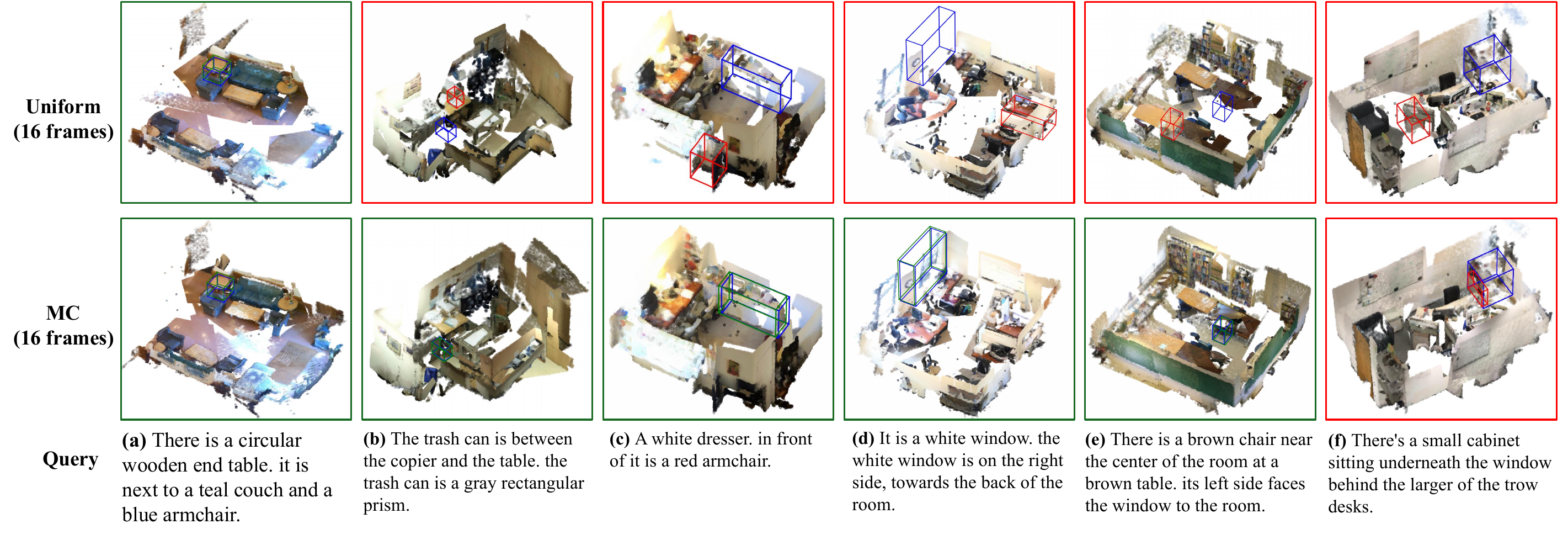} 
\vspace{-10pt}
\caption{
The visualization results on ScanRefer. 
The green/red/blue colors indicate the correct/incorrect/ground truth boxes.
}
\vspace{-10pt}
\label{fig:visulization}

\end{figure*}

\input{table/ablation_ghead}

\input{table/ablation_video_modeling}

\begin{figure}[t] 
\centering 
\includegraphics[width=0.46\textwidth]{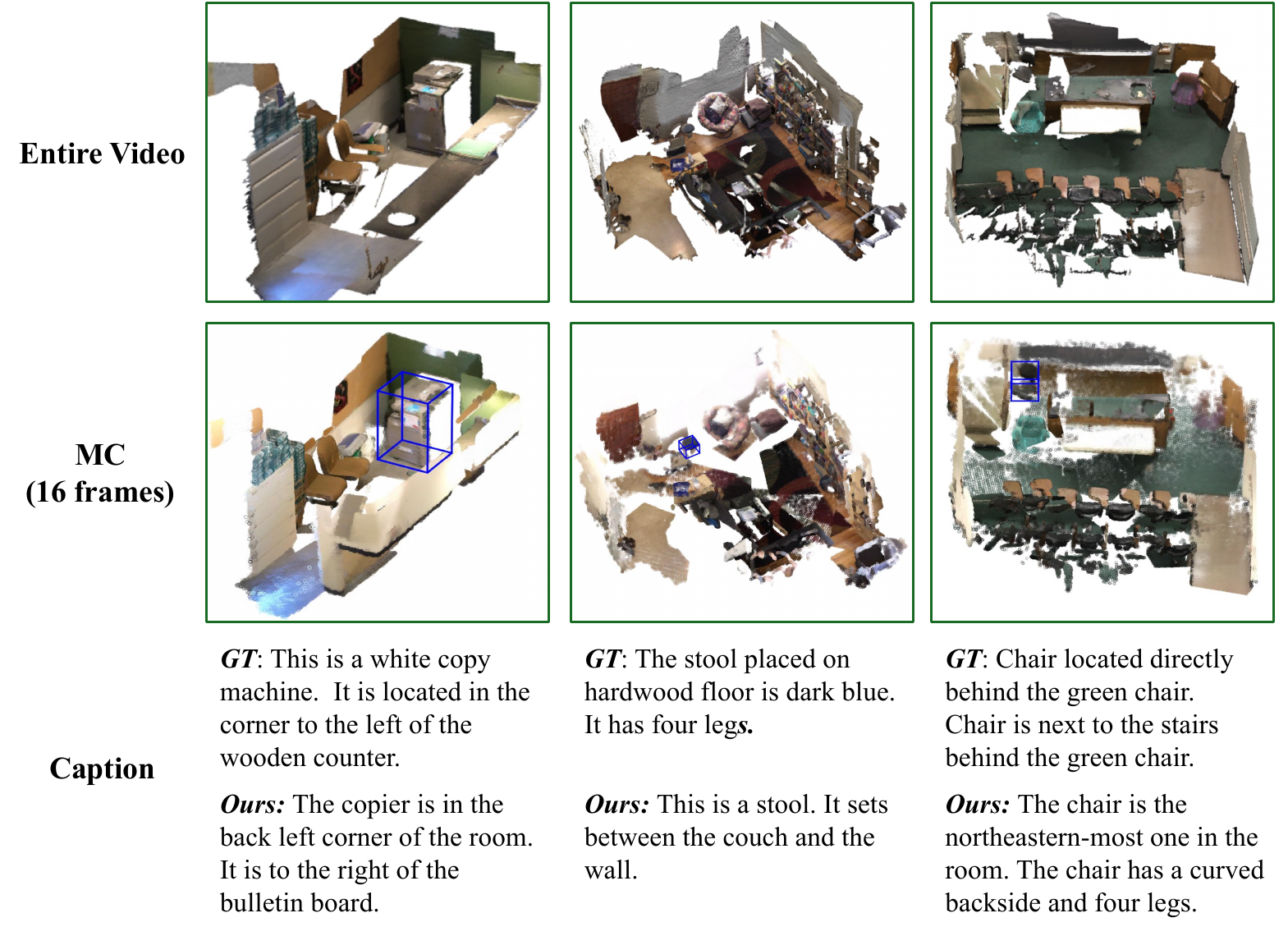} 
\vspace{-5pt}
\caption{
The visualization results on Scan2Cap. 
The input boxes are marked in blue.
}
\vspace{-10pt}

\label{fig:visualization_caption}

\end{figure}

\vspace{3pt}
\noindent \textbf{Effectiveness of 3D-PE.} Table~\ref{table:ablation_pe} presents the performance using different 3D position encoding (3D-PE) and coordinate aggregation strategies. The model was trained on the entire multi-task dataset, with 32 frames under the uniform frame sampling.
We first assess the impact of different 3D-PE using average coordinate aggregation. 
\red{ 
The introduction of 3D-PE leads to a consistent performance increase in 3D dense captioning and 3D visual grounding in all aggregation variants.}
Specifically, for dense captioning tasks, which require locating objects within video sequences based on query bounding boxes, the absence of 3D-PE results in significantly degraded performance. 
Additionally, $\textsl{MLP}$ position encoding proves more effective for grounding tasks, while $\textsl{Sin}$ PE works better for others.
We then evaluate different coordinate aggregation strategies using $\textsl{Sin}$ position encoding. Since the 3D coordinates are more complex and variable than those in 2D images, simply using the coordinates at the center of a 2D patch doesn't accurately reflect the spatial location. As shown in the result, using average 3D coordinates better represents the 3D location within a patch, while using the minimum and maximum 3D coordinates within the patch is also a viable alternative.

\vspace{3pt}
\noindent \textbf{Ablation for 3D Visual Grounding.}
As illustrated in Table \ref{table:ablation_ghead}, we investigate the effect of varying patch sizes of object embeddings and loss functions.
Since LLaVA-Video \cite{llavavideo} downsamples the image patches again before feeding them into LLM, we can generate object embeddings with patch sizes of 14 or 27.
For ScanRefer, switching the patch size from 27 to 14 leads to improved accuracy, with Acc@0.5 from 48.93\% to 50.08\%. 
This improvement may be attributed to the smaller patch size to capture more precise object features.
Additionally, since we model the grounding task by leveraging the similarity between object embeddings and the \(\langle \text{ground} \rangle\) token, the use of BCE loss may impose overly strict constraints. Replacing BCE loss with InfoNCE loss consistently improves performance.

\vspace{3pt}
\noindent \textbf{Effectiveness of Representing 3D Scenes as Videos.}
To eliminate the influence of the backbone, we conducted ablation experiments by comparing our 3D-as-video paradigm with voxel modeling based on the same backbone, LLaVA-Video.
Following LLaVA-3D, we average the patch features corresponding to the same voxel to obtain the voxel features, and sample 3,096 of these features as input for the LLM.
As shown in Table \ref{table:ablation_video_modeling}, there is a notable improvement of \textbf{28.9} C@0.5 on Scan2Cap, from 83.8 to 54.9.
Note that our dataset is only 26\% of LLaVA-3D, resulting the lower results compared to the LLaVA-3D.
For ScanQA, which relies more heavily on visual perception, our approach still achieves a 2\% gain in accuracy compared to voxel modeling.
This indicates that our approach leverages spatiotemporal priors specifically, rather than merely benefiting from stronger visual features.

\vspace{3pt}
\noindent \textbf{Visualization.}
Figure \ref{fig:visulization} presents visualization results on the validation split of ScanRefer.
The first and second rows indicate the rendered point clouds using uniform sampling and maximum coverage sampling with 16 frames.
We observe that maximum coverage sampling usually provides more complete scenes than uniform sampling.
For complex cases like (b-e), uniform sampling tends to miss smaller or peripheral objects, leading to prediction failures.
Figure \ref{fig:visualization_caption} presents visualization results on the Scan2Cap validation set. The first two rows show the rendered point clouds of the full video and the frames sampled using the MC strategy. With only 16 frames, nearly the entire scene's information is captured. Meanwhile,  using the proposed 3D-PE, the model accurately describes the specified target object.

%% file: table/ablation_frame.tex
\begin{table*}[]{

\centering
\renewcommand\arraystretch{1}
\setlength{\tabcolsep}{2.4mm}{
\resizebox{0.94\textwidth}{!}{
\begin{tabular}{cc|c|ccccccccc} \toprule
\multirow{2}{*}{\makecell[c]{Frame\\Number}} & \multirow{2}{*}{\makecell[c]{Sampling\\Strategy}}  & \multirow{2}{*}{\makecell[c]{Inference\\Time}}  & \multicolumn{2}{c}{ScanRefer} & \multicolumn{2}{c}{Multi3dRefer}& \multicolumn{2}{c}{Scan2Cap}  & \multicolumn{2}{c}{ScanQA} & \multicolumn{1}{c}{SQA3D}\\
 & & & Acc@0.25 & Acc@0.5 & F1@0.25 & F1@0.5 & B-4@0.5 & C@0.5 & C & EM & EM \\ \midrule
  \multicolumn{5}{l}{\textit{\textbf{Fixed Frame Number}}}\\  
  \hline
\multirow{2}{*}{8} & Uniform & \multirow{2}{*}{309ms}  & 48.93 & 43.50 & 49.80 & 45.40 & 37.34 & 68.82 & 94.98 & 27.57 & 56.77 \\
 & MC & & 53.47 & 47.41 & 53.55 & 48.54 & 38.77 & 73.08 & 96.37 & 28.00 & 56.97 \\ \hline
\multirow{2}{*}{16} & Uniform & \multirow{2}{*}{537ms} & 55.42 & 49.17 & 54.95 & 49.82 & 39.39 & 76.96 & 99.86 & 28.96 & 57.70 \\
 & MC & & 56.46 & 50.11 & 56.65 & 51.39 & 39.59 & 76.84 & 100.63 & 29.49 & 57.82\\ \hline
\multirow{2}{*}{32} & Uniform & \multirow{2}{*}{1050ms} & 58.11 & \textbf{51.72} & \textbf{58.02} & \textbf{52.68} & \textbf{41.30} & \textbf{83.76} & 102.06 & 30.09 & 58.56 \\
 & MC & & \textbf{58.27} & 51.68 & 57.93 & 52.50 & 40.32 & 81.58 & \textbf{102.33} & \textbf{30.35} & \textbf{59.25} \\
 \midrule
\multicolumn{5}{l}{\textit{\textbf{Adaptive Frame Number}}}\\  
\hline
$\approx$18 & MC$^{*}$ & 527ms & 57.86 & 51.18 & 57.87 & 52.40 & 40.18 & 80.00 & 100.54 & 29.50 & 57.72 \\
\midrule
\multicolumn{5}{l}{\textit{\textbf{Previous SOTA}}}\\  
\hline
\multicolumn{2}{c|}{LLaVA-3D~\cite{llava3d}} & 433ms & 54.1 & 42.4 & -- & -- & 41.1 & 79.2 & 91.7 & 27.0 & 55.6 \\
\bottomrule
\end{tabular}}
}
\vspace{-5pt}
\caption{Ablation study for the effect of frame sampling strategy. ``MC''  represents maximum coverage sampling. ``MC$^{*}$'' denotes sampling frames until over 95\% of the scene's voxels are covered or a maximum of 32 frames is reached.}
\vspace{-7pt}
\label{table:ablation_frame}
}
\end{table*}

%% file: table/ablation_pe.tex





\begin{table}[]{
\centering
\resizebox{0.96\linewidth}{!}{
\begin{tabular}{c|c|lcc|c} \toprule
\multirow{2}{*}{\textbf{3D-PE}} & \multirow{2}{*}{\textbf{Coord.}}   &Scan2Cap& \multicolumn{2}{c|}{ScanRefer}  & \multicolumn{1}{c}{ScanQA} \\
&  &C@0.5 & Acc@0.25 & Acc@0.5 & EM \\ \midrule

None & \multirow{3}{*}{Avg}  &31.03 & 57.50 & 50.84 & 30.03 \\
MLP &   &76.23 & \textbf{59.63} & \textbf{52.98} & 29.62 \\
\underline{Sin} &   &\textbf{83.76} & 58.11 & 51.72 & \textbf{30.09} \\ 

\midrule

\multirow{3}{*}{Sin} & Center  &80.88 & 57.53 & 51.06 & 29.39 \\
& Min-Max  &82.75 & 58.05 & \textbf{51.77} & \textbf{30.18}  \\
& \underline{Avg}  &\textbf{83.76} & \textbf{58.11} & 51.72 & 30.09  \\ 

\bottomrule
\end{tabular}}
\vspace{-5pt}
\caption{Ablation study for the effect of coordinate encoding. ``Coord.'' means the method for aggregating the coordinates.
}
\label{table:ablation_pe}
}
\vspace{-5pt}
\end{table}

%% file: table/ablation_ghead.tex

\begin{table}[]{
\centering
\resizebox{0.96\linewidth}{!}{
\begin{tabular}{c|c|cc|cc} \toprule
\multirow{2}{*}{\makecell[c]{\textbf{Patch}\\\textbf{Size}}} & \multirow{2}{*}{\textbf{Loss}} & \multicolumn{2}{c|}{ScanRefer} & \multicolumn{2}{c}{Multi3DRefer} \\
& & Acc@0.25 & Acc@0.5 & Acc@0.5 & Acc@0.5 \\ \midrule
14 & InfoNCE & \textbf{56.44} & \textbf{50.08} & \textbf{56.31} & \textbf{51.05} \\
27 & InfoNCE & 55.23 & 48.93 & 56.13 & 50.90   \\
14& BCE  &  51.63 & 45.82 & 46.07 & 41.47  \\
\bottomrule
\end{tabular}}
\vspace{-5pt}
\caption{Ablation study for the effect of visual grounding.
We train the model separately on the ScanRefer and Multi3DRefer datasets.}
\vspace{-0pt}
\label{table:ablation_ghead}
}
\end{table}

%% file: table/ablation_video_modeling.tex
\begin{table}[]{
\centering
\resizebox{0.98\linewidth}{!}{
\begin{tabular}{c|cc|cc|c} \toprule
\multirow{1}{*}{\textbf{}} & \multicolumn{2}{c|}{Scan2Cap} & \multicolumn{2}{c|}{ScanRefer} & \multicolumn{1}{c}{ScanQA} \\
& C@0.5 & B@0.5 & Acc@0.25 & Acc@0.5 & EM \\ \midrule
Voxel & 54.9 & 34.3 & 56.1 & 49.8 & 28.1 \\
Video & 83.8 (+28.9) & 41.3 (+7.0) & 58.1 (+2.0)& 51.7 (+1.9) & 30.1 (+2.0) \\

\bottomrule
\end{tabular}}
\vspace{-5px}
\caption{Ablation study for the 3D scene representation.
}
\vspace{-10px}
\label{table:ablation_video_modeling}
}
\end{table}

%% file: sec/5_conclusion.tex
\section{Conclusion}
In this paper, we propose a novel paradigm to effectively exploit Video LLMs for 3D scene understanding.
It incorporates 3D position encoding into video representations and employs a carefully-designed multi-task training recipe, thereby facilitating a suite of 3D scene understanding tasks.
We further introduce a maximum coverage sampling strategy to optimize the trade-off between computational costs and model performance.
Our extensive experimental results demonstrate the superiority of our method.

\section*{Acknowledgements}
This work is supported by National Key R\&D Program of China (Project No. 2022ZD0161200, 2022ZD0161201). This work is also supported by Hong Kong Research Grant Council - Early Career Scheme (Grant No. 24200223).

%% file: sec/X_appendix.tex
\section{Implementation Details}
\vspace{5pt}
\noindent \textbf{Dataset Statistics.}
We present the detailed statistics for training and testing data in Table \ref{tab:data_stat1} and \ref{tab:data_stat2}, respectively.
Following previous work \cite{leo, chatscene}, we adopt the validation set for ScanRefer \cite{scanrefer}, Multi3DRefer \cite{multi3drefer}, Scan2Cap \cite{scan2cap}, ScanQA \cite{scan2cap}, and the test set for SQA3D \cite{sqa3d}.
All data have been converted to LLaVA \cite{llava} format, and we conduct statistics in this format.

\vspace{5pt}
\noindent \textbf{Evaluation Details.}
For ScanRefer \cite{scanrefer}, we select the object with the highest similarity as the prediction.
For Multi3DRefer \cite{multi3drefer}, we select objects with the highest probabilities such that their cumulative probability exceeds a given threshold p, which is empirically set to 0.25.
For Scan2Cap \cite{scan2cap}, we follow \cite{leo} to evaluate the captioning performance by inserting ``sos'' and ``eos'' at the start and end of the prediction, respectively.
Responses are generated using greedy sampling for 3D dense captioning and 3D question answering tasks.

\section{Detailed Comparison}

\vspace{5pt}
\noindent \textbf{SQA3D.}
We conduct a detailed evaluation on the test split of the SQA3D \cite{sqa3d} dataset.
As shown in Table \ref{tab:sqa3d}, our model achieves the best performance on all categories of questions with an average EM at 58.86\%, outperforming the previous state-of-the-art method LLaVA-3D \cite{llava3d} by 2.94\% on the average EM.

\vspace{5pt}
\noindent \textbf{Scan2Cap.} As shown in Table \ref{tab:scan2cap}, we provide a detailed comparison on the validation set of Scan2Cap \cite{scan2cap}.
During inference, we utilize the object proposals from \cite{leo}, which include 50 predicted objects extracted with Mask3D \cite{mask3d} for each scan.
From the table, we can see our method achieves state-of-the-art results on CIDEr and BLEU-4 at 83.77 and 42.43, respectively.

\vspace{5pt}
\noindent \textbf{ScanRefer.} 
We present a detailed comparison for ScanRefer \cite{scanrefer} in Table \ref{tab:scanrefer}. The table shows that our method reaches a peak of 58.12\% Acc@0.25 and 51.72\% Acc@0.5, surpassing ChatScene \cite{chatscene} by 2.6\% and 1.5\%, respectively.

\vspace{5pt}
\noindent \textbf{Multi3DRefer.} 
We follow previous work \cite{multi3drefer} to report the metrics across all question types, where ``ZT'' denotes zero-target, ``ST'' denotes single-target, ``MT'' denotes multi-target, ``w/ D'' and ``w/o D'' denote `with and without distractors, respectively.
As shown in Table \ref{tab:multi3drefer}, our method outperforms previous methods on ``ZT w/o D'', ``ZT w/ D'', and ``ST w/D'' types.
However, the performance for ``MT'' is lower than ChatScene \cite{chatscene}, suggesting that our method still struggles to distinguish similar objects.

\input{table/data_stat}
\input{table/sqa3d}
\input{table/scan2cap}
\input{table/scanrefer}
\input{table/multi3drefer}
\input{table/scanqa}

\vspace{5pt}
\noindent \textbf{ScanQA.} 
We test our model on the validation set of ScanQA \cite{scanqa}.
Compared to previous top-tier models, our \model{} achieves a relative improvement of 10.7\% and 11.9\% on EM@1 and CIDEr, respectively.

%% file: table/data_stat.tex
\begin{table}[]
\setlength{\tabcolsep}{2.8mm}{
\resizebox{\linewidth}{!}{
\begin{tabular}{lcccc}
\toprule
& \multirow{2}{*}{\shortstack{Data \\Count}} & \multirow{2}{*}{\shortstack{Scan \\Count}} & \multirow{2}{*}{\shortstack{Ques \\length}} & \multirow{2}{*}{\shortstack{Answer \\Length}} \\
& \\ \bottomrule
ScanRefer \cite{scanrefer} & 36,665 & 562 & 24.9 & -- \\
Multi3DRefer \cite{multi3drefer} & 43,838 & 562 & 34.8 & --\\
Scan2Cap \cite{scan2cap} & 36,665 & 562 & 13.0 & 17.9 \\
ScanQA \cite{scanqa} & 26,515 & 562 & 13.7 & 2.4 \\
SQA3D \cite{sqa3d} & 79,445 & 518 & 37.8 & 1.1 \\

\bottomrule
\end{tabular}}}
\caption{
Detailed statistics for training data.
We report the average lengths for questions and answers, respectively.
}
\label{tab:data_stat1}
\end{table}

\begin{table}[]
\setlength{\tabcolsep}{2.5mm}{
\resizebox{\linewidth}{!}{
\begin{tabular}{lcccc}
\toprule
& \multirow{2}{*}{\shortstack{Data \\Count}} & \multirow{2}{*}{\shortstack{Scan \\Count}} & \multirow{2}{*}{\shortstack{Ques \\length}} & \multirow{2}{*}{\shortstack{Answer \\Length}} \\
& \\ \bottomrule
ScanRefer \cite{scanrefer} (Val) & 9,508 & 141 & 25.0 & -- \\
Multi3DRefer \cite{multi3drefer} (Val) & 11,120 & 141 & 34.7 & --\\
Scan2Cap \cite{scan2cap} (Val) & 2,068 & 141 & 13.0 & 18.7 \\
ScanQA \cite{scanqa} (Val) & 4,675 & 71 & 13.8 & 2.4 \\
SQA3D \cite{sqa3d} (Test) & 3,519 & 67 & 36.3 & 1.1 \\

\bottomrule
\end{tabular}}}
\caption{Detailed statistics for testing data.
We report the average lengths for questions and answers, respectively.
}
\label{tab:data_stat2}
\end{table}

%% file: table/sqa3d.tex

\begin{table}[]
\setlength{\tabcolsep}{1mm}{
\resizebox{\linewidth}{!}{
\begin{tabular}{lccccccc}
\toprule
\multirow{2}{*}{Method} & \multicolumn{6}{c}{Test set} & \multirow{2}{*}{\textbf{Avg.}} \\ \cmidrule{2-7}
 & What & Is & How & Can & Which & Others &  \\
\midrule
SQA3D~\cite{sqa3d} & 31.6 & 63.8 & 46.0 & 69.5 & 43.9 & 45.3 & 46.6 \\
3D-VisTA~\cite{3dvista} & 34.8 & 63.3 & 45.4 & 69.8 & 47.2 & 48.1 & 48.5 \\
LLaVA-Video\cite{llavavideo} & 42.7 & 56.3 & 47.5 & 55.3 & 50.1 & 47.2 & 48.5 \\
Scene-LLM~\cite{scenellm} & 40.9 & {69.1} & 45.0 & {70.8} & 47.2 & 52.3 & 54.2 \\
LEO \cite{leo} & -- & -- & -- & -- & -- & -- & 50.0 \\
ChatScene \cite{chatscene} & 45.4 & 67.0 & 52.0 & 69.5 & 49.9 & 55.0 & 54.6 \\
LLaVA-3D \cite{llava3d} & -- & -- & -- & -- & -- & -- & 55.6 \\
\rowcolor{lightgray} \model{} (Uniform) & 51.1 & 72.4 & 55.5 & 69.8 & 51.3 & 56.0 & \textbf{58.6} \\
\rowcolor{lightgray} \model{} (MC) & 50.0 & 70.7 & 57.9 & 69.8 & 50.1 & 55.8 & 57.7 \\
\bottomrule
\end{tabular}}}
\caption{Performance comparison on the test set of SQA3D \cite{sqa3d}.}
\label{tab:sqa3d}
\end{table}

%% file: table/scan2cap.tex
\begin{table}[]
\centering
\resizebox{1\linewidth}{!}{
\begin{tabular}{lcccccccc}
\toprule
\multirow{2}{*}{Method} & \multicolumn{4}{c}{@0.5} \\
 & C & B-4 & M & R \\
\midrule
Scan2Cap~\cite{scan2cap} & 39.08 & 23.32 & 21.97 & 44.48 \\
3DJCG~\cite{3djcg}& 49.48 & 31.03 & 24.22 & 50.80 \\
D3Net~\cite{d3net}  & 62.64 & 35.68 & 25.72 & 53.90 \\
3D-VisTA \cite{3dvista}  & 66.9 & 34.0 & 27.1 & 54.3 \\
LL3DA \cite{ll3da}  & 65.19 & 36.79 & 25.97 & 55.06 \\
LEO \cite{leo} & 68.4 & 36.9 & 27.7 & 57.8 \\
{ChatScene} \cite{chatscene} & {77.19} & 36.34 & {28.01} & {58.12} \\
LLaVA-3D \cite{llava3d} & 79.21  &41.12 & \textbf{30.21} & \textbf{63.41} \\
\rowcolor{lightgray} \model{} (Uniform) & \textbf{83.77} & \textbf{42.43} & 28.87 & 62.34 \\
\rowcolor{lightgray} \model{} (MC) & 80.00 & 40.18 & 28.49 & 61.68 \\
\bottomrule
\end{tabular}}
\caption{{Performance comparison on the validation set of Scan2Cap~\cite{scan2cap}. C, B-4, M, R represent CIDEr, BLEU-4, Meteor, Rouge-L, respectively.}} \label{tab:scan2cap}
\end{table}

%% file: table/scanrefer.tex
\begin{table*}[!h]
\centering
\resizebox{1.0\linewidth}{!}{
\setlength{\tabcolsep}{3.5mm}{
\begin{tabular}{lccccccc}
\toprule
\multirow{2}{*}{Method} & \multirow{2}{*}{Venue} & \multicolumn{2}{c}{Unique} & \multicolumn{2}{c}{Multiple} & \multicolumn{2}{c}{Overall} \\
 &  & Acc@0.25 & Acc@0.5 & Acc@0.25 & Acc@0.5 & Acc@0.25 & Acc@0.5 \\
 \midrule
ScanRefer~\cite{scanrefer} & ECCV20 & 76.33 & 53.51 & 32.73 & 21.11 & 41.19 & 27.40 \\
MVT~\cite{mvt} & CVPR22 & 77.67 & 66.45 & 31.92 & 25.26 & 40.80 & 33.26 \\
3DVG-Transformer~\cite{3dvg-trans} & ICCV21 & 81.93 & 60.64 & 39.30 & 28.42 & 47.57 & 34.67 \\
ViL3DRel~\cite{vil3drel} & NeurIPS22 & 81.58 & 68.62 & 40.30 & 30.71 & 47.94 & 37.73 \\
3DJCG~\cite{3djcg} & CVPR22 & 83.47 & 64.34 & 41.39 & 30.82 & 49.56 & 37.33 \\
D3Net~\cite{d3net} & ECCV22 & -- & 72.04 & -- & 30.05 & -- & 37.87 \\
M3DRef-CLIP~\cite{multi3drefer} & ICCV23 & 85.3 & 77.2 & 43.8 & 36.8 & 51.9 & 44.7 \\
3D-VisTA~\cite{3dvista} & ICCV23 & 81.6 & 75.1 & 43.7 & 39.1 & 50.6 & 45.8 \\
3D-LLM (Flamingo) \cite{3d-llm} & NeurIPS23 & -- & -- & -- & -- & 21.2 & -- \\
3D-LLM (BLIP2-flant5) \cite{3d-llm} & NeurIPS23 & -- & -- & -- & -- & 30.3 & -- \\
Grounded 3D-LLM \cite{grounded-3dllm} & ArXiv24 & -- & -- & -- & -- & 47.9 & 44.1 \\
PQ3D \cite{pq3d} & ECCV24 & 86.7 & 78.3 & 51.5 & 46.2 & 57.0 & 51.2 \\
ChatScene \cite{chatscene} & NeurIPS24 & 89.59 & 82.49 & 47.78 & 42.90 & 55.52 & 50.23 \\
LLaVA-3D \cite{llava3d} & ArXiv24 & -- & -- & -- & -- & 54.1 & 42.2\\
\rowcolor{lightgray} \model{} (Uniform) & -- & 87.97 & 78.32 & 50.93 & 45.32 & \textbf{58.12} & \textbf{51.72}  \\
\rowcolor{lightgray} \model{} (MC) & -- & 86.61 & 77.02 & 50.95 & 44.96 & 57.87 & 51.18 \\ 
\bottomrule
\end{tabular}}}
\caption{Performance comparison on the validation set of ScanRefer \cite{scanrefer}.
``Unique'' and ``Multiple'' depends on whether there are other objects of the same class as the target object.
} \label{tab:scanrefer}
\end{table*}

%% file: table/multi3drefer.tex
\begin{table*}[!h]{
\centering
\resizebox{1.0\linewidth}{!}{
\begin{tabular}{l|cc|cccc|cc|cc}
\toprule
\multirow{2}{*}{Method}  & ZT w/o D & ZT w/ D & \multicolumn{2}{c}{ST w/o D} & \multicolumn{2}{c}{ST w/ D} &\multicolumn{2}{c}{MT} & \multicolumn{2}{c}{ALL} \\
 & F1 & F1 & F1@0.25 & F1@0.5 & F1@0.25 & F1@0.5 & F1@0.25 & F1@0.5 & F1@0.25 & F1@0.5 \\
 \midrule
M3DRef-CLIP~\cite{multi3drefer} & 81.8 & 39.4 & 53.5 & 47.8 & 34.6 & 30.6 & 43.6 & 37.9 & 42.8 & 38.4 \\
D3Net~\cite{d3net}  & 81.6 & 32.5 & -- & 38.6 & -- & 23.3 & -- & 35.0 & -- & 32.2 \\
3DJCG~\cite{3djcg} & {94.1} & {66.9} & -- & 26.0 & -- & 16.7 & -- & 26.2 & -- & 26.6 \\
Grounded 3D-LLM \cite{grounded-3dllm} & -- & -- & -- & -- & -- & -- & -- & -- & 45.2 & 40.6 \\
PQ3D \cite{pq3d} & 85.4 & 57.7 & -- & 68.5 & -- & 43.6 & -- & 40.9 & -- & 50.1 \\
ChatScene \cite{chatscene}  & 90.3 & 62.6 & {82.9} & {75.9} & {49.1} & {44.5} & {45.7} & {41.1} & {57.1} & {52.4} \\
\rowcolor{lightgray} \model{} (Uniform) & 94.7 & 78.5 & 82.6 & 73.4 & 52.1 & 47.2 & 40.8 & 35.7 & \textbf{58.0} & \textbf{52.7} \\
\rowcolor{lightgray} \model{} (MC) & 94.1 & 76.7 & 81.2 & 72.6 & 52.7 & 47.4 & 40.6 & 35.3 & 57.9 & 52.4 \\
\bottomrule
\end{tabular}}
\caption{{Performance comparison on the validation set of Multi3DRefer \cite{multi3drefer}. ZT: zero-target, ST: single-target, MT: multi-target, D: distractor.}} \label{tab:multi3drefer}
}
\end{table*}

%% file: table/scanqa.tex
\begin{table*}[!h]
\centering
\resizebox{1.0\linewidth}{!}{
\setlength{\tabcolsep}{3.5mm}{
\begin{tabular}{lccccccccc}
\toprule
Method & Venue & EM & B-1 & B-2 & B-3 & B-4 & ROUGE-L & METEOR & CIDEr \\
\midrule
ScanQA~\cite{scanqa} & CVPR22 & 21.05 & 30.24 & 20.40 & 15.11 & 10.08 & 33.33 & 13.14 & 64.86  \\
3D-VisTA \cite{3dvista} & ICCV23 & 22.4 & -- & -- & -- & 10.4 & 35.7 & 13.9 & 69.6 \\
Oryx-34B \cite{oryx} & ArXiv24 & -- & 38.0 & 24.6 & -- & -- & 37.3 & 15.0 & 72.3 \\
LLaVA-Video-7B \cite{llavavideo} & ArXiv24 & -- & 39.71 & 26.57 & 9.33 & 3.09 & 44.62 & 17.72 & 88.70 \\
3D-LLM (Flamingo) \cite{3d-llm} & NeurIPS23 & 20.4 & 30.3 & 17.8 & 12.0 & 7.2 & 32.3 & 12.2 & 59.2 \\
3D-LLM (BLIP2-flant5)~\cite{3d-llm} & NeurIPS23 & 20.5 & 39.3 & 25.2 & 18.4 & 12.0 & 35.7 & 14.5 & 69.4  \\
Chat-3D \cite{chat3d} & ArXiv23 & -- & 29.1 & -- & -- & 6.4 & 28.5 & 11.9 & 53.2 \\
NaviLLM \cite{navillm} & CVPR24 & 23.0 & -- & -- & -- & 12.5 & 38.4 & 15.4 & 75.9 \\
LL3DA~\cite{ll3da} & CVPR24 & -- & -- & -- & \multicolumn{1}{l}{--} & 13.53 & 37.31 & 15.88 & 76.79 \\
Scene-LLM~\cite{scenellm} & ArXiv24 & 27.2 & 43.6 & 26.8 & 19.1 & 12.0 & 40.0 & 16.6 & 80.0 \\
LEO~\cite{leo} & ICML24 & -- & -- & -- & -- & 11.5 & 39.3 & 16.2 & 80.0 \\
Grounded 3D-LLM \cite{grounded-3dllm} & ArXiv24 & -- & -- & -- & -- & 13.4 & -- & -- & 72.7\\
ChatScene \cite{chatscene} & NeurIPS24 & 21.62 & 43.20 & 29.06 & 20.57 & 14.31 & 41.56 & 18.00 & 87.70 \\
LLaVA-3D \cite{llava3d} & arXiv24 & 27.0 & -- & -- & -- & 14.5 & \textbf{50.1} & \textbf{20.7} &  91.7 \\
\rowcolor{lightgray} \model{} (Uniform) & -- & \textbf{30.10} & \textbf{47.05} & \textbf{31.70} & \textbf{22.83} & 16.17 & 49.02 & 19.84 & \textbf{102.06} \\
\rowcolor{lightgray} \model{} (MC) & -- & 29.50 & 46.23 & 31.22 & 22.71 & \textbf{16.28} & 48.19 & 19.36 & 100.54 \\
\bottomrule
\end{tabular}}}
\caption{Performance comparison on the validation set of ScanQA \cite{scanqa}.
EM indicates exact match accuracy, and B-1, B-2, B-3, B-4 denote BLEU-1, -2, -3, -4, respectively.
} \label{tab:scanqa}
\end{table*}

%% file: main.bbl
\begin{thebibliography}{54}
\providecommand{\natexlab}[1]{#1}
\providecommand{\url}[1]{\texttt{#1}}
\expandafter\ifx\csname urlstyle\endcsname\relax
  \providecommand{\doi}[1]{doi: #1}\else
  \providecommand{\doi}{doi: \begingroup \urlstyle{rm}\Url}\fi

\bibitem[Anil et~al.(2023)Anil, Borgeaud, Wu, Alayrac, Yu, Soricut, Schalkwyk, Dai, Hauth, Millican, Silver, Petrov, Johnson, Antonoglou, Schrittwieser, Glaese, Chen, Pitler, Lillicrap, Lazaridou, Firat, Molloy, Isard, Barham, Hennigan, Lee, Viola, Reynolds, Xu, Doherty, Collins, Meyer, Rutherford, Moreira, Ayoub, Goel, Tucker, Piqueras, Krikun, Barr, Savinov, Danihelka, Roelofs, White, Andreassen, von Glehn, Yagati, Kazemi, Gonzalez, Khalman, Sygnowski, and et~al.]{gemini}
Rohan Anil, Sebastian Borgeaud, Yonghui Wu, Jean{-}Baptiste Alayrac, Jiahui Yu, Radu Soricut, Johan Schalkwyk, Andrew~M. Dai, Anja Hauth, Katie Millican, David Silver, Slav Petrov, Melvin Johnson, Ioannis Antonoglou, Julian Schrittwieser, Amelia Glaese, Jilin Chen, Emily Pitler, Timothy~P. Lillicrap, Angeliki Lazaridou, Orhan Firat, James Molloy, Michael Isard, Paul~Ronald Barham, Tom Hennigan, Benjamin Lee, Fabio Viola, Malcolm Reynolds, Yuanzhong Xu, Ryan Doherty, Eli Collins, Clemens Meyer, Eliza Rutherford, Erica Moreira, Kareem Ayoub, Megha Goel, George Tucker, Enrique Piqueras, Maxim Krikun, Iain Barr, Nikolay Savinov, Ivo Danihelka, Becca Roelofs, Ana{\"{\i}}s White, Anders Andreassen, Tamara von Glehn, Lakshman Yagati, Mehran Kazemi, Lucas Gonzalez, Misha Khalman, Jakub Sygnowski, and et al.
\newblock Gemini: {A} family of highly capable multimodal models.
\newblock \emph{ArXiv preprint}, abs/2312.11805, 2023.

\bibitem[Azuma et~al.(2022)Azuma, Miyanishi, Kurita, and Kawanabe]{scanqa}
Daichi Azuma, Taiki Miyanishi, Shuhei Kurita, and Motoaki Kawanabe.
\newblock Scanqa: 3d question answering for spatial scene understanding.
\newblock In \emph{{IEEE/CVF} Conference on Computer Vision and Pattern Recognition, {CVPR} 2022, New Orleans, LA, USA, June 18-24, 2022}, pages 19107--19117. {IEEE}, 2022.

\bibitem[Cai et~al.(2022)Cai, Zhao, Zhang, Sheng, and Xu]{3djcg}
Daigang Cai, Lichen Zhao, Jing Zhang, Lu Sheng, and Dong Xu.
\newblock 3djcg: {A} unified framework for joint dense captioning and visual grounding on 3d point clouds.
\newblock In \emph{{IEEE/CVF} Conference on Computer Vision and Pattern Recognition, {CVPR} 2022, New Orleans, LA, USA, June 18-24, 2022}, pages 16443--16452. {IEEE}, 2022.

\bibitem[Chen et~al.(2024{\natexlab{a}})Chen, Xu, Kirmani, Ichter, Sadigh, Guibas, and Xia]{spatial-vlm}
Boyuan Chen, Zhuo Xu, Sean Kirmani, Brian Ichter, Dorsa Sadigh, Leonidas~J. Guibas, and Fei Xia.
\newblock Spatialvlm: Endowing vision-language models with spatial reasoning capabilities.
\newblock In \emph{{IEEE/CVF} Conference on Computer Vision and Pattern Recognition, {CVPR} 2024, Seattle, WA, USA, June 16-22, 2024}, pages 14455--14465. {IEEE}, 2024{\natexlab{a}}.

\bibitem[Chen et~al.(2020)Chen, Chang, and Nie{\ss}ner]{scanrefer}
Dave~Zhenyu Chen, Angel~X. Chang, and Matthias Nie{\ss}ner.
\newblock Scanrefer: 3d object localization in {RGB-D} scans using natural language.
\newblock In \emph{Computer Vision - {ECCV} 2020 - 16th European Conference, Glasgow, UK, August 23-28, 2020, Proceedings, Part {XX}}, pages 202--221. Springer, 2020.

\bibitem[Chen et~al.(2021)Chen, Gholami, Nie{\ss}ner, and Chang]{scan2cap}
Dave~Zhenyu Chen, Ali Gholami, Matthias Nie{\ss}ner, and Angel~X. Chang.
\newblock Scan2cap: Context-aware dense captioning in {RGB-D} scans.
\newblock In \emph{{IEEE} Conference on Computer Vision and Pattern Recognition, {CVPR} 2021, virtual, June 19-25, 2021}, pages 3193--3203. Computer Vision Foundation / {IEEE}, 2021.

\bibitem[Chen et~al.(2022{\natexlab{a}})Chen, Wu, Nie{\ss}ner, and Chang]{d3net}
Dave~Zhenyu Chen, Qirui Wu, Matthias Nie{\ss}ner, and Angel~X. Chang.
\newblock D\({}^{\mbox{3}}\)net: {A} unified speaker-listener architecture for 3d dense captioning and visual grounding.
\newblock In \emph{Computer Vision - {ECCV} 2022 - 17th European Conference, Tel Aviv, Israel, October 23-27, 2022, Proceedings, Part {XXXII}}, pages 487--505. Springer, 2022{\natexlab{a}}.

\bibitem[Chen et~al.(2022{\natexlab{b}})Chen, Guhur, Tapaswi, Schmid, and Laptev]{vil3drel}
Shizhe Chen, Pierre{-}Louis Guhur, Makarand Tapaswi, Cordelia Schmid, and Ivan Laptev.
\newblock Language conditioned spatial relation reasoning for 3d object grounding.
\newblock In \emph{Advances in Neural Information Processing Systems 35: Annual Conference on Neural Information Processing Systems 2022, NeurIPS 2022, New Orleans, LA, USA, November 28 - December 9, 2022}, 2022{\natexlab{b}}.

\bibitem[Chen et~al.(2024{\natexlab{b}})Chen, Chen, Zhang, Li, Yu, Fei, Zhu, Fan, and Chen]{ll3da}
Sijin Chen, Xin Chen, Chi Zhang, Mingsheng Li, Gang Yu, Hao Fei, Hongyuan Zhu, Jiayuan Fan, and Tao Chen.
\newblock {LL3DA:} visual interactive instruction tuning for omni-3d understanding, reasoning, and planning.
\newblock In \emph{{IEEE/CVF} Conference on Computer Vision and Pattern Recognition, {CVPR} 2024, Seattle, WA, USA, June 16-22, 2024}, pages 26418--26428. {IEEE}, 2024{\natexlab{b}}.

\bibitem[Chen et~al.(2024{\natexlab{c}})Chen, Yang, Huang, Wang, Lyu, Xu, Lin, and Pang]{grounded-3dllm}
Yilun Chen, Shuai Yang, Haifeng Huang, Tai Wang, Ruiyuan Lyu, Runsen Xu, Dahua Lin, and Jiangmiao Pang.
\newblock Grounded 3d-llm with referent tokens.
\newblock \emph{ArXiv preprint}, abs/2405.10370, 2024{\natexlab{c}}.

\bibitem[Chen et~al.(2024{\natexlab{d}})Chen, Wu, Wang, Su, Chen, Xing, Zhong, Zhang, Zhu, Lu, Li, Luo, Lu, Qiao, and Dai]{internvl}
Zhe Chen, Jiannan Wu, Wenhai Wang, Weijie Su, Guo Chen, Sen Xing, Muyan Zhong, Qinglong Zhang, Xizhou Zhu, Lewei Lu, Bin Li, Ping Luo, Tong Lu, Yu Qiao, and Jifeng Dai.
\newblock Intern {VL:} scaling up vision foundation models and aligning for generic visual-linguistic tasks.
\newblock In \emph{{IEEE/CVF} Conference on Computer Vision and Pattern Recognition, {CVPR} 2024, Seattle, WA, USA, June 16-22, 2024}, pages 24185--24198. {IEEE}, 2024{\natexlab{d}}.

\bibitem[Dai et~al.(2017)Dai, Chang, Savva, Halber, Funkhouser, and Nie{\ss}ner]{scannet}
Angela Dai, Angel~X. Chang, Manolis Savva, Maciej Halber, Thomas~A. Funkhouser, and Matthias Nie{\ss}ner.
\newblock Scannet: Richly-annotated 3d reconstructions of indoor scenes.
\newblock In \emph{2017 {IEEE} Conference on Computer Vision and Pattern Recognition, {CVPR} 2017, Honolulu, HI, USA, July 21-26, 2017}, pages 2432--2443. {IEEE} Computer Society, 2017.

\bibitem[Dosovitskiy et~al.(2021)Dosovitskiy, Beyer, Kolesnikov, Weissenborn, Zhai, Unterthiner, Dehghani, Minderer, Heigold, Gelly, Uszkoreit, and Houlsby]{vit}
Alexey Dosovitskiy, Lucas Beyer, Alexander Kolesnikov, Dirk Weissenborn, Xiaohua Zhai, Thomas Unterthiner, Mostafa Dehghani, Matthias Minderer, Georg Heigold, Sylvain Gelly, Jakob Uszkoreit, and Neil Houlsby.
\newblock An image is worth 16x16 words: Transformers for image recognition at scale.
\newblock In \emph{9th International Conference on Learning Representations, {ICLR} 2021, Virtual Event, Austria, May 3-7, 2021}. OpenReview.net, 2021.

\bibitem[Dubey et~al.(2024)Dubey, Jauhri, Pandey, Kadian, Al{-}Dahle, Letman, Mathur, Schelten, Yang, Fan, Goyal, Hartshorn, Yang, Mitra, Sravankumar, Korenev, Hinsvark, Rao, Zhang, Rodriguez, Gregerson, Spataru, Rozi{\`{e}}re, Biron, Tang, Chern, Caucheteux, Nayak, Bi, Marra, McConnell, Keller, Touret, Wu, Wong, Ferrer, Nikolaidis, Allonsius, Song, Pintz, Livshits, Esiobu, Choudhary, Mahajan, Garcia{-}Olano, Perino, Hupkes, Lakomkin, AlBadawy, Lobanova, Dinan, Smith, Radenovic, Zhang, Synnaeve, Lee, Anderson, Nail, Mialon, Pang, Cucurell, Nguyen, Korevaar, Xu, Touvron, Zarov, Ibarra, Kloumann, Misra, Evtimov, Copet, Lee, Geffert, Vranes, Park, Mahadeokar, Shah, van~der Linde, Billock, Hong, Lee, Fu, Chi, Huang, Liu, Wang, Yu, Bitton, Spisak, Park, Rocca, Johnstun, Saxe, Jia, Alwala, Upasani, Plawiak, Li, Heafield, Stone, and et~al.]{llama3}
Abhimanyu Dubey, Abhinav Jauhri, Abhinav Pandey, Abhishek Kadian, Ahmad Al{-}Dahle, Aiesha Letman, Akhil Mathur, Alan Schelten, Amy Yang, Angela Fan, Anirudh Goyal, Anthony Hartshorn, Aobo Yang, Archi Mitra, Archie Sravankumar, Artem Korenev, Arthur Hinsvark, Arun Rao, Aston Zhang, Aur{\'{e}}lien Rodriguez, Austen Gregerson, Ava Spataru, Baptiste Rozi{\`{e}}re, Bethany Biron, Binh Tang, Bobbie Chern, Charlotte Caucheteux, Chaya Nayak, Chloe Bi, Chris Marra, Chris McConnell, Christian Keller, Christophe Touret, Chunyang Wu, Corinne Wong, Cristian~Canton Ferrer, Cyrus Nikolaidis, Damien Allonsius, Daniel Song, Danielle Pintz, Danny Livshits, David Esiobu, Dhruv Choudhary, Dhruv Mahajan, Diego Garcia{-}Olano, Diego Perino, Dieuwke Hupkes, Egor Lakomkin, Ehab AlBadawy, Elina Lobanova, Emily Dinan, Eric~Michael Smith, Filip Radenovic, Frank Zhang, Gabriel Synnaeve, Gabrielle Lee, Georgia~Lewis Anderson, Graeme Nail, Gr{\'{e}}goire Mialon, Guan Pang, Guillem Cucurell, Hailey Nguyen, Hannah Korevaar, Hu Xu, Hugo
  Touvron, Iliyan Zarov, Imanol~Arrieta Ibarra, Isabel~M. Kloumann, Ishan Misra, Ivan Evtimov, Jade Copet, Jaewon Lee, Jan Geffert, Jana Vranes, Jason Park, Jay Mahadeokar, Jeet Shah, Jelmer van~der Linde, Jennifer Billock, Jenny Hong, Jenya Lee, Jeremy Fu, Jianfeng Chi, Jianyu Huang, Jiawen Liu, Jie Wang, Jiecao Yu, Joanna Bitton, Joe Spisak, Jongsoo Park, Joseph Rocca, Joshua Johnstun, Joshua Saxe, Junteng Jia, Kalyan~Vasuden Alwala, Kartikeya Upasani, Kate Plawiak, Ke Li, Kenneth Heafield, Kevin Stone, and et al.
\newblock The llama 3 herd of models.
\newblock \emph{ArXiv preprint}, abs/2407.21783, 2024.

\bibitem[Fu et~al.(2024)Fu, Liu, Chen, Nie, and Xiong]{scenellm}
Rao Fu, Jingyu Liu, Xilun Chen, Yixin Nie, and Wenhan Xiong.
\newblock Scene-llm: Extending language model for 3d visual understanding and reasoning.
\newblock \emph{ArXiv preprint}, abs/2403.11401, 2024.

\bibitem[Hong et~al.(2023)Hong, Zhen, Chen, Zheng, Du, Chen, and Gan]{3d-llm}
Yining Hong, Haoyu Zhen, Peihao Chen, Shuhong Zheng, Yilun Du, Zhenfang Chen, and Chuang Gan.
\newblock 3d-llm: Injecting the 3d world into large language models.
\newblock In \emph{Advances in Neural Information Processing Systems 36: Annual Conference on Neural Information Processing Systems 2023, NeurIPS 2023, New Orleans, LA, USA, December 10 - 16, 2023}, 2023.

\bibitem[Huang et~al.(2023)Huang, Wang, Huang, Liu, Cheng, Zhao, Jin, and Zhao]{chatscene}
Haifeng Huang, Zehan Wang, Rongjie Huang, Luping Liu, Xize Cheng, Yang Zhao, Tao Jin, and Zhou Zhao.
\newblock Chat-3d v2: Bridging 3d scene and large language models with object identifiers.
\newblock \emph{ArXiv preprint}, abs/2312.08168, 2023.

\bibitem[Huang et~al.(2024)Huang, Yong, Ma, Linghu, Li, Wang, Li, Zhu, Jia, and Huang]{leo}
Jiangyong Huang, Silong Yong, Xiaojian Ma, Xiongkun Linghu, Puhao Li, Yan Wang, Qing Li, Song{-}Chun Zhu, Baoxiong Jia, and Siyuan Huang.
\newblock An embodied generalist agent in 3d world.
\newblock In \emph{Forty-first International Conference on Machine Learning, {ICML} 2024, Vienna, Austria, July 21-27, 2024}. OpenReview.net, 2024.

\bibitem[Huang et~al.(2022)Huang, Chen, Jia, and Wang]{mvt}
Shijia Huang, Yilun Chen, Jiaya Jia, and Liwei Wang.
\newblock Multi-view transformer for 3d visual grounding.
\newblock In \emph{{IEEE/CVF} Conference on Computer Vision and Pattern Recognition, {CVPR} 2022, New Orleans, LA, USA, June 18-24, 2022}, pages 15503--15512. {IEEE}, 2022.

\bibitem[Kang et~al.(2024)Kang, Huang, Shang, Shah, and Yan]{robin3d}
Weitai Kang, Haifeng Huang, Yuzhang Shang, Mubarak Shah, and Yan Yan.
\newblock Robin3d: Improving 3d large language model via robust instruction tuning.
\newblock \emph{ArXiv preprint}, abs/2410.00255, 2024.

\bibitem[Khuller et~al.(1999)Khuller, Moss, and Naor]{KHULLER199939}
Samir Khuller, Anna Moss, and Joseph~(Seffi) Naor.
\newblock The budgeted maximum coverage problem.
\newblock \emph{Information Processing Letters}, 70\penalty0 (1):\penalty0 39--45, 1999.

\bibitem[Kim et~al.(2024)Kim, Pertsch, Karamcheti, Xiao, Balakrishna, Nair, Rafailov, Foster, Lam, Sanketi, Vuong, Kollar, Burchfiel, Tedrake, Sadigh, Levine, Liang, and Finn]{openvla}
Moo~Jin Kim, Karl Pertsch, Siddharth Karamcheti, Ted Xiao, Ashwin Balakrishna, Suraj Nair, Rafael Rafailov, Ethan~Paul Foster, Grace Lam, Pannag Sanketi, Quan Vuong, Thomas Kollar, Benjamin Burchfiel, Russ Tedrake, Dorsa Sadigh, Sergey Levine, Percy Liang, and Chelsea Finn.
\newblock Openvla: An open-source vision-language-action model.
\newblock \emph{ArXiv preprint}, abs/2406.09246, 2024.

\bibitem[Li et~al.(2024{\natexlab{a}})Li, Zhang, Guo, Zhang, Li, Zhang, Zhang, Li, Liu, and Li]{llava-onevision}
Bo Li, Yuanhan Zhang, Dong Guo, Renrui Zhang, Feng Li, Hao Zhang, Kaichen Zhang, Yanwei Li, Ziwei Liu, and Chunyuan Li.
\newblock Llava-onevision: Easy visual task transfer.
\newblock \emph{ArXiv preprint}, abs/2408.03326, 2024{\natexlab{a}}.

\bibitem[Li et~al.(2023)Li, Li, Savarese, and Hoi]{blip2}
Junnan Li, Dongxu Li, Silvio Savarese, and Steven C.~H. Hoi.
\newblock {BLIP-2:} bootstrapping language-image pre-training with frozen image encoders and large language models.
\newblock In \emph{International Conference on Machine Learning, {ICML} 2023, 23-29 July 2023, Honolulu, Hawaii, {USA}}, pages 19730--19742. {PMLR}, 2023.

\bibitem[Li et~al.(2024{\natexlab{b}})Li, Wang, He, Li, Wang, Liu, Wang, Xu, Chen, Lou, Wang, and Qiao]{mvbench}
Kunchang Li, Yali Wang, Yinan He, Yizhuo Li, Yi Wang, Yi Liu, Zun Wang, Jilan Xu, Guo Chen, Ping Lou, Limin Wang, and Yu Qiao.
\newblock Mvbench: {A} comprehensive multi-modal video understanding benchmark.
\newblock In \emph{{IEEE/CVF} Conference on Computer Vision and Pattern Recognition, {CVPR} 2024, Seattle, WA, USA, June 16-22, 2024}, pages 22195--22206. {IEEE}, 2024{\natexlab{b}}.

\bibitem[Lin et~al.(2024)Lin, Yin, Ping, Molchanov, Shoeybi, and Han]{vila}
Ji Lin, Hongxu Yin, Wei Ping, Pavlo Molchanov, Mohammad Shoeybi, and Song Han.
\newblock {VILA:} on pre-training for visual language models.
\newblock In \emph{{IEEE/CVF} Conference on Computer Vision and Pattern Recognition, {CVPR} 2024, Seattle, WA, USA, June 16-22, 2024}, pages 26679--26689. {IEEE}, 2024.

\bibitem[Linghu et~al.(2024)Linghu, Huang, Niu, Ma, Jia, and Huang]{msqa}
Xiongkun Linghu, Jiangyong Huang, Xuesong Niu, Xiaojian~(Shawn) Ma, Baoxiong Jia, and Siyuan Huang.
\newblock Multi-modal situated reasoning in 3d scenes.
\newblock In \emph{Advances in Neural Information Processing Systems 38: Annual Conference on Neural Information Processing Systems 2024, NeurIPS 2024, Vancouver, BC, Canada, December 10 - 15, 2024}, 2024.

\bibitem[Liu et~al.(2024{\natexlab{a}})Liu, Dong, Wang, Rao, Tang, Ma, and Krishna]{coarse-correspondence}
Benlin Liu, Yuhao Dong, Yiqin Wang, Yongming Rao, Yansong Tang, Wei{-}Chiu Ma, and Ranjay Krishna.
\newblock Coarse correspondence elicit 3d spacetime understanding in multimodal language model.
\newblock \emph{ArXiv preprint}, abs/2408.00754, 2024{\natexlab{a}}.

\bibitem[Liu et~al.(2023)Liu, Li, Wu, and Lee]{llava}
Haotian Liu, Chunyuan Li, Qingyang Wu, and Yong~Jae Lee.
\newblock Visual instruction tuning.
\newblock In \emph{Advances in Neural Information Processing Systems 36: Annual Conference on Neural Information Processing Systems 2023, NeurIPS 2023, New Orleans, LA, USA, December 10 - 16, 2023}, 2023.

\bibitem[Liu et~al.(2024{\natexlab{b}})Liu, Dong, Liu, Hu, Lu, and Rao]{oryx}
Zuyan Liu, Yuhao Dong, Ziwei Liu, Winston Hu, Jiwen Lu, and Yongming Rao.
\newblock Oryx {MLLM:} on-demand spatial-temporal understanding at arbitrary resolution.
\newblock \emph{ArXiv preprint}, abs/2409.12961, 2024{\natexlab{b}}.

\bibitem[Ma et~al.(2023)Ma, Yong, Zheng, Li, Liang, Zhu, and Huang]{sqa3d}
Xiaojian Ma, Silong Yong, Zilong Zheng, Qing Li, Yitao Liang, Song{-}Chun Zhu, and Siyuan Huang.
\newblock {SQA3D:} situated question answering in 3d scenes.
\newblock In \emph{The Eleventh International Conference on Learning Representations, {ICLR} 2023, Kigali, Rwanda, May 1-5, 2023}. OpenReview.net, 2023.

\bibitem[Majumdar et~al.(2024)Majumdar, Ajay, Zhang, Putta, Yenamandra, Henaff, Silwal, McVay, Maksymets, Arnaud, Yadav, Li, Newman, Sharma, Berges, Zhang, Agrawal, Bisk, Batra, Kalakrishnan, Meier, Paxton, Sax, and Rajeswaran]{openeqa}
Arjun Majumdar, Anurag Ajay, Xiaohan Zhang, Pranav Putta, Sriram Yenamandra, Mikael Henaff, Sneha Silwal, Paul McVay, Oleksandr Maksymets, Sergio Arnaud, Karmesh Yadav, Qiyang Li, Ben Newman, Mohit Sharma, Vincent{-}Pierre Berges, Shiqi Zhang, Pulkit Agrawal, Yonatan Bisk, Dhruv Batra, Mrinal Kalakrishnan, Franziska Meier, Chris Paxton, Alexander Sax, and Aravind Rajeswaran.
\newblock Openeqa: Embodied question answering in the era of foundation models.
\newblock In \emph{{IEEE/CVF} Conference on Computer Vision and Pattern Recognition, {CVPR} 2024, Seattle, WA, USA, June 16-22, 2024}, pages 16488--16498. {IEEE}, 2024.

\bibitem[OpenAI(2023)]{gpt-4}
OpenAI.
\newblock {GPT-4} technical report.
\newblock \emph{ArXiv preprint}, abs/2303.08774, 2023.

\bibitem[Schult et~al.(2023)Schult, Engelmann, Hermans, Litany, Tang, and Leibe]{mask3d}
Jonas Schult, Francis Engelmann, Alexander Hermans, Or Litany, Siyu Tang, and Bastian Leibe.
\newblock Mask3d: Mask transformer for 3d semantic instance segmentation.
\newblock In \emph{{IEEE} International Conference on Robotics and Automation, {ICRA} 2023, London, UK, May 29 - June 2, 2023}, pages 8216--8223. {IEEE}, 2023.

\bibitem[van~den Oord et~al.(2018)van~den Oord, Li, and Vinyals]{DBLP:journals/corr/abs-1807-03748}
A{\"{a}}ron van~den Oord, Yazhe Li, and Oriol Vinyals.
\newblock Representation learning with contrastive predictive coding.
\newblock \emph{ArXiv preprint}, abs/1807.03748, 2018.

\bibitem[Vaswani et~al.(2017)Vaswani, Shazeer, Parmar, Uszkoreit, Jones, Gomez, Kaiser, and Polosukhin]{DBLP:conf/nips/VaswaniSPUJGKP17}
Ashish Vaswani, Noam Shazeer, Niki Parmar, Jakob Uszkoreit, Llion Jones, Aidan~N. Gomez, Lukasz Kaiser, and Illia Polosukhin.
\newblock Attention is all you need.
\newblock In \emph{Advances in Neural Information Processing Systems 30: Annual Conference on Neural Information Processing Systems 2017, December 4-9, 2017, Long Beach, CA, {USA}}, pages 5998--6008, 2017.

\bibitem[Vedantam et~al.(2015)Vedantam, Zitnick, and Parikh]{cider}
Ramakrishna Vedantam, C.~Lawrence Zitnick, and Devi Parikh.
\newblock Cider: Consensus-based image description evaluation.
\newblock In \emph{{IEEE} Conference on Computer Vision and Pattern Recognition, {CVPR} 2015, Boston, MA, USA, June 7-12, 2015}, pages 4566--4575. {IEEE} Computer Society, 2015.

\bibitem[Wald et~al.(2019)Wald, Avetisyan, Navab, Tombari, and Nie{\ss}ner]{3rscan}
Johanna Wald, Armen Avetisyan, Nassir Navab, Federico Tombari, and Matthias Nie{\ss}ner.
\newblock {RIO:} 3d object instance re-localization in changing indoor environments.
\newblock In \emph{2019 {IEEE/CVF} International Conference on Computer Vision, {ICCV} 2019, Seoul, Korea (South), October 27 - November 2, 2019}, pages 7657--7666. {IEEE}, 2019.

\bibitem[Wang et~al.(2024)Wang, Bai, Tan, Wang, Fan, Bai, Chen, Liu, Wang, Ge, Fan, Dang, Du, Ren, Men, Liu, Zhou, Zhou, and Lin]{qwen2vl}
Peng Wang, Shuai Bai, Sinan Tan, Shijie Wang, Zhihao Fan, Jinze Bai, Keqin Chen, Xuejing Liu, Jialin Wang, Wenbin Ge, Yang Fan, Kai Dang, Mengfei Du, Xuancheng Ren, Rui Men, Dayiheng Liu, Chang Zhou, Jingren Zhou, and Junyang Lin.
\newblock Qwen2-vl: Enhancing vision-language model's perception of the world at any resolution.
\newblock \emph{ArXiv preprint}, abs/2409.12191, 2024.

\bibitem[Wang et~al.(2023)Wang, Huang, Zhao, Zhang, and Zhao]{chat3d}
Zehan Wang, Haifeng Huang, Yang Zhao, Ziang Zhang, and Zhou Zhao.
\newblock Chat-3d: Data-efficiently tuning large language model for universal dialogue of 3d scenes.
\newblock \emph{ArXiv preprint}, abs/2308.08769, 2023.

\bibitem[Wu et~al.(2024)Wu, Jing, Cheang, Chen, Xu, Li, Liu, Li, and Kong]{gr1}
Hongtao Wu, Ya Jing, Chilam Cheang, Guangzeng Chen, Jiafeng Xu, Xinghang Li, Minghuan Liu, Hang Li, and Tao Kong.
\newblock Unleashing large-scale video generative pre-training for visual robot manipulation.
\newblock In \emph{The Twelfth International Conference on Learning Representations, {ICLR} 2024, Vienna, Austria, May 7-11, 2024}. OpenReview.net, 2024.

\bibitem[Xu et~al.(2024)Xu, Wang, Wang, Chen, Pang, and Lin]{pointllm}
Runsen Xu, Xiaolong Wang, Tai Wang, Yilun Chen, Jiangmiao Pang, and Dahua Lin.
\newblock Pointllm: Empowering large language models to understand point clouds.
\newblock In \emph{Computer Vision - {ECCV} 2024 - 18th European Conference, Milan, Italy, September 29-October 4, 2024, Proceedings, Part {XXV}}, pages 131--147. Springer, 2024.

\bibitem[Yang et~al.(2024)Yang, Yang, Hui, Zheng, Yu, Zhou, Li, Li, Liu, Huang, Dong, Wei, Lin, Tang, Wang, Yang, Tu, Zhang, Ma, Yang, Xu, Zhou, Bai, He, Lin, Dang, Lu, Chen, Yang, Li, Xue, Ni, Zhang, Wang, Peng, Men, Gao, Lin, Wang, Bai, Tan, Zhu, Li, Liu, Ge, Deng, Zhou, Ren, Zhang, Wei, Ren, Liu, Fan, Yao, Zhang, Wan, Chu, Liu, Cui, Zhang, Guo, and Fan]{qwen2}
An Yang, Baosong Yang, Binyuan Hui, Bo Zheng, Bowen Yu, Chang Zhou, Chengpeng Li, Chengyuan Li, Dayiheng Liu, Fei Huang, Guanting Dong, Haoran Wei, Huan Lin, Jialong Tang, Jialin Wang, Jian Yang, Jianhong Tu, Jianwei Zhang, Jianxin Ma, Jianxin Yang, Jin Xu, Jingren Zhou, Jinze Bai, Jinzheng He, Junyang Lin, Kai Dang, Keming Lu, Keqin Chen, Kexin Yang, Mei Li, Mingfeng Xue, Na Ni, Pei Zhang, Peng Wang, Ru Peng, Rui Men, Ruize Gao, Runji Lin, Shijie Wang, Shuai Bai, Sinan Tan, Tianhang Zhu, Tianhao Li, Tianyu Liu, Wenbin Ge, Xiaodong Deng, Xiaohuan Zhou, Xingzhang Ren, Xinyu Zhang, Xipin Wei, Xuancheng Ren, Xuejing Liu, Yang Fan, Yang Yao, Yichang Zhang, Yu Wan, Yunfei Chu, Yuqiong Liu, Zeyu Cui, Zhenru Zhang, Zhifang Guo, and Zhihao Fan.
\newblock Qwen2 technical report.
\newblock \emph{ArXiv preprint}, abs/2407.10671, 2024.

\bibitem[Yeshwanth et~al.(2023)Yeshwanth, Liu, Nie{\ss}ner, and Dai]{scannetpp}
Chandan Yeshwanth, Yueh{-}Cheng Liu, Matthias Nie{\ss}ner, and Angela Dai.
\newblock Scannet++: {A} high-fidelity dataset of 3d indoor scenes.
\newblock In \emph{{IEEE/CVF} International Conference on Computer Vision, {ICCV} 2023, Paris, France, October 1-6, 2023}, pages 12--22. {IEEE}, 2023.

\bibitem[Zhang et~al.(2023{\natexlab{a}})Zhang, Li, and Bing]{videollama}
Hang Zhang, Xin Li, and Lidong Bing.
\newblock Video-{LL}a{MA}: An instruction-tuned audio-visual language model for video understanding.
\newblock In \emph{Proceedings of the 2023 Conference on Empirical Methods in Natural Language Processing: System Demonstrations}, pages 543--553, Singapore, 2023{\natexlab{a}}. Association for Computational Linguistics.

\bibitem[Zhang et~al.(2023{\natexlab{b}})Zhang, Gong, and Chang]{multi3drefer}
Yiming Zhang, ZeMing Gong, and Angel~X. Chang.
\newblock Multi3drefer: Grounding text description to multiple 3d objects.
\newblock In \emph{{IEEE/CVF} International Conference on Computer Vision, {ICCV} 2023, Paris, France, October 1-6, 2023}, page 15179. {IEEE}, 2023{\natexlab{b}}.

\bibitem[Zhang et~al.(2024)Zhang, Wu, Li, Li, Ma, Liu, and Li]{llavavideo}
Yuanhan Zhang, Jinming Wu, Wei Li, Bo Li, Zejun Ma, Ziwei Liu, and Chunyuan Li.
\newblock Video instruction tuning with synthetic data, 2024.

\bibitem[Zhao et~al.(2021)Zhao, Cai, Sheng, and Xu]{3dvg-trans}
Lichen Zhao, Daigang Cai, Lu Sheng, and Dong Xu.
\newblock 3dvg-transformer: Relation modeling for visual grounding on point clouds.
\newblock In \emph{2021 {IEEE/CVF} International Conference on Computer Vision, {ICCV} 2021, Montreal, QC, Canada, October 10-17, 2021}, pages 2908--2917. {IEEE}, 2021.

\bibitem[Zheng et~al.(2024)Zheng, Huang, Zhao, Zhong, and Wang]{navillm}
Duo Zheng, Shijia Huang, Lin Zhao, Yiwu Zhong, and Liwei Wang.
\newblock Towards learning a generalist model for embodied navigation.
\newblock In \emph{{IEEE/CVF} Conference on Computer Vision and Pattern Recognition, {CVPR} 2024, Seattle, WA, USA, June 16-22, 2024}, pages 13624--13634. {IEEE}, 2024.

\bibitem[Zhu et~al.(2024{\natexlab{a}})Zhu, Wang, Zhang, Chen, and Liu]{scanreason}
Chenming Zhu, Tai Wang, Wenwei Zhang, Kai Chen, and Xihui Liu.
\newblock Scanreason: Empowering 3d visual grounding with reasoning capabilities.
\newblock In \emph{Computer Vision - {ECCV} 2024 - 18th European Conference, Milan, Italy, September 29-October 4, 2024, Proceedings, Part {VIII}}, pages 151--168. Springer, 2024{\natexlab{a}}.

\bibitem[Zhu et~al.(2024{\natexlab{b}})Zhu, Wang, Zhang, Pang, and Liu]{llava3d}
Chenming Zhu, Tai Wang, Wenwei Zhang, Jiangmiao Pang, and Xihui Liu.
\newblock Llava-3d: {A} simple yet effective pathway to empowering lmms with 3d-awareness.
\newblock \emph{ArXiv preprint}, abs/2409.18125, 2024{\natexlab{b}}.

\bibitem[Zhu et~al.(2023)Zhu, Ma, Chen, Deng, Huang, and Li]{3dvista}
Ziyu Zhu, Xiaojian Ma, Yixin Chen, Zhidong Deng, Siyuan Huang, and Qing Li.
\newblock 3d-vista: Pre-trained transformer for 3d vision and text alignment.
\newblock In \emph{{IEEE/CVF} International Conference on Computer Vision, {ICCV} 2023, Paris, France, October 1-6, 2023}, pages 2899--2909. {IEEE}, 2023.

\bibitem[Zhu et~al.(2024{\natexlab{c}})Zhu, Zhang, Ma, Niu, Chen, Jia, Deng, Huang, and Li]{pq3d}
Ziyu Zhu, Zhuofan Zhang, Xiaojian Ma, Xuesong Niu, Yixin Chen, Baoxiong Jia, Zhidong Deng, Siyuan Huang, and Qing Li.
\newblock Unifying 3d vision-language understanding via promptable queries.
\newblock In \emph{Computer Vision - {ECCV} 2024 - 18th European Conference, Milan, Italy, September 29-October 4, 2024, Proceedings, Part {XLIV}}, pages 188--206. Springer, 2024{\natexlab{c}}.

\bibitem[Zitkovich et~al.(2023)Zitkovich, Yu, Xu, Xu, Xiao, Xia, Wu, Wohlhart, Welker, Wahid, Vuong, Vanhoucke, Tran, Soricut, Singh, Singh, Sermanet, Sanketi, Salazar, Ryoo, Reymann, Rao, Pertsch, Mordatch, Michalewski, Lu, Levine, Lee, Lee, Leal, Kuang, Kalashnikov, Julian, Joshi, Irpan, Ichter, Hsu, Herzog, Hausman, Gopalakrishnan, Fu, Florence, Finn, Dubey, Driess, Ding, Choromanski, Chen, Chebotar, Carbajal, Brown, Brohan, Arenas, and Han]{rt2}
Brianna Zitkovich, Tianhe Yu, Sichun Xu, Peng Xu, Ted Xiao, Fei Xia, Jialin Wu, Paul Wohlhart, Stefan Welker, Ayzaan Wahid, Quan Vuong, Vincent Vanhoucke, Huong~T. Tran, Radu Soricut, Anikait Singh, Jaspiar Singh, Pierre Sermanet, Pannag~R. Sanketi, Grecia Salazar, Michael~S. Ryoo, Krista Reymann, Kanishka Rao, Karl Pertsch, Igor Mordatch, Henryk Michalewski, Yao Lu, Sergey Levine, Lisa Lee, Tsang{-}Wei~Edward Lee, Isabel Leal, Yuheng Kuang, Dmitry Kalashnikov, Ryan Julian, Nikhil~J. Joshi, Alex Irpan, Brian Ichter, Jasmine Hsu, Alexander Herzog, Karol Hausman, Keerthana Gopalakrishnan, Chuyuan Fu, Pete Florence, Chelsea Finn, Kumar~Avinava Dubey, Danny Driess, Tianli Ding, Krzysztof~Marcin Choromanski, Xi Chen, Yevgen Chebotar, Justice Carbajal, Noah Brown, Anthony Brohan, Montserrat~Gonzalez Arenas, and Kehang Han.
\newblock {RT-2:} vision-language-action models transfer web knowledge to robotic control.
\newblock In \emph{Conference on Robot Learning, CoRL 2023, 6-9 November 2023, Atlanta, GA, {USA}}, pages 2165--2183. {PMLR}, 2023.

\end{thebibliography}
